\documentclass{article}

\PassOptionsToPackage{numbers,compress,sort}{natbib}
\usepackage[preprint]{neurips_2026}

\usepackage[utf8]{inputenc}
\usepackage[T1]{fontenc}
\usepackage{url}
\usepackage{booktabs}
\usepackage{amsfonts}
\usepackage{amsmath}
\usepackage{amssymb}
\usepackage{nicefrac}
\usepackage{microtype}
\usepackage{xcolor}
\usepackage{enumitem}
\usepackage{hyperref}

\usepackage{graphicx}
\usepackage{caption}
\usepackage{float}
\usepackage{wrapfig}
\usepackage{multirow}
\usepackage{pifont}
\usepackage[ruled,linesnumbered,vlined]{algorithm2e}
\usepackage{xspace}

\providecommand{\appendices}{\appendix}

\newcommand{\name}{TrackFish3D\xspace}

\makeatletter
\renewcommand{\@noticestring}{%
	40th Conference on Neural Information Processing Systems (NeurIPS 2026).
}
\makeatother

\title{TrackFish3D: Self-Supervised 3D Tracking of\\Schooling Fish from Multi-view Videos}

\author{%
{\bfseries Patt Phurtivilai$^{1}$,
Zhiyang Dou$^{1,\dagger}$,
Yifan Wu$^{1}$,
Kinfung Chu$^{4}$} \\[0.35em]
{\bfseries Yuan Liu$^{2}$,
Lei Yang$^{1}$,
Wenping Wang$^{3}$,
Taku Komura$^{1,\dagger}$} \\[0.6em]
{\mdseries $^{1}$ The University of Hong Kong, Hong Kong, China} \\[0.2em]
{\mdseries $^{2}$ Hong Kong University of Science and Technology, Hong Kong, China} \\[0.2em]
{\mdseries $^{3}$ Texas A\&M University, College Station, TX, US} \\[0.2em]
{\mdseries $^{4}$ Centre for Transformative Garment Production}
}

\date{}

\begin{document}

\maketitle

\begingroup
\renewcommand{\thefootnote}{\fnsymbol{footnote}}
\footnotetext[2]{Corresponding authors.}
\endgroup

\begin{abstract}
Quantifying collective fish behavior requires accurate trajectories, yet multi-view 3D tracking remains challenging due to frequent occlusions, visually similar individuals, and the long-standing scarcity of identity annotations. We present \name, a geometry-driven self-supervised framework for dense multi-camera 3D tracking of schooling fish. Instead of relying on appearance-based re-identification or manually annotated identities, \name turns calibrated multi-view geometry into supervision: triangulation and reprojection consistency provide pseudo-associations, while a geometric encoder and global association transformer learn all-to-all cross-view correspondence within each frame. To make these associations identity-aware, \name introduces a self-supervised contrastive objective that separates co-visible individuals in the embedding space, together with a temporal predictor that preserves identities and bridges short occlusions across frames. The resulting model is trained once on unlabeled footage and applied directly to unseen test videos, requiring no cross-view identity labels, temporal annotations, 3D ground truth, appearance features, or test-time optimization. On our benchmark, 
\name improves 3D Multi-Object Tracking Accuracy from 87.7\% for the strongest baseline to 95.8\%. On the 3D-ZeF zebrafish benchmark, it achieves 81.1\% MOTA, compared with 77.4\% for the best geometric baseline. \name also generalizes beyond fish, achieving strong results on real-world bird tracking. \href{https://pattgene.github.io/trackfish3d-projectpage/}{\textcolor{blue}{Project page}}.
\vspace{-3mm}
\end{abstract}

\section{Introduction}

\begin{wrapfigure}{r}{0.2\columnwidth}
    \vspace{-12pt}
    \centering
    \includegraphics[width=0.2\columnwidth]{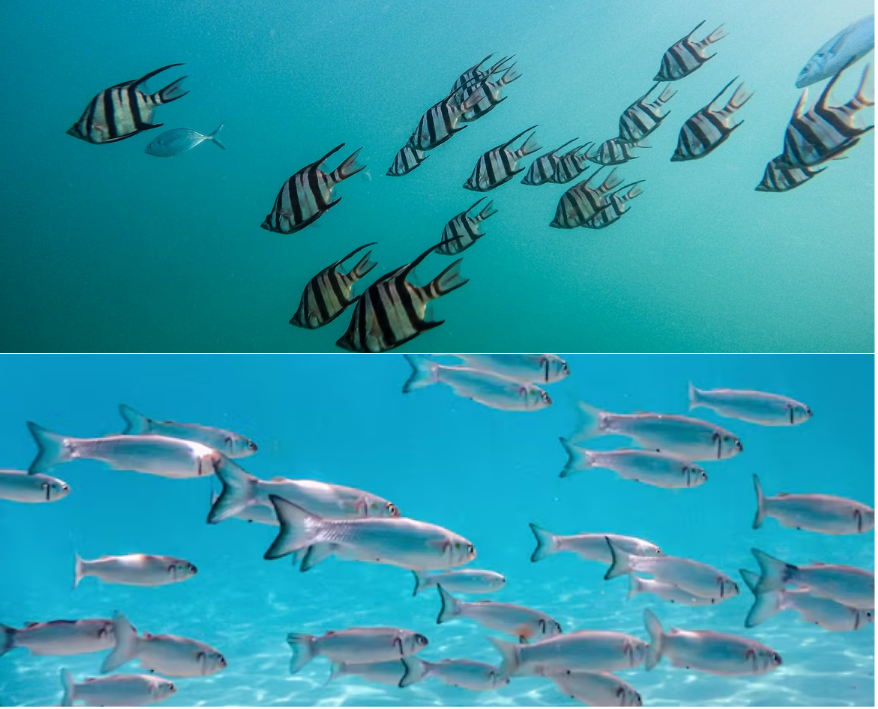}
    \caption{Dense fish school in the wild share a nearly identical appearance with mutual occlusion.\protect\footnotemark }
    \label{fig:schoolfish}
    \vspace{-10pt}
\end{wrapfigure}
\footnotetext{Image sources: Unsplash (\url{unsplash.com}); Freepik (\url{freepik.com}).}

Collective animal behaviour, from fish schools to bird flocks and insect swarms, emerges from local interaction rules and is central to behavioural ecology and computational biology~\cite{vicsek2012collective,cavagna2018physics,couzin2002collective,reynolds1987flocks,aoki1982simulation}. 
Understanding these rules informs models of decision-making, predator evasion, and swarm robotics~\cite{couzin2005effective,heins2024collective,herbertread2011inferring,newbolt2019flow,chung2018survey,zhou2022swarm}. Among them, \textit{schooling fish} provide a tractable system because their group structure and interactions vary with density, group size, and environmental context~\cite{couzin2002collective,couzin2005effective,calovi2015collective,jiang2023collective,wu2024cbil}. 
Yet naturalistic 3D schooling remains incompletely understood, requiring dense 3D trajectories for every individual over time~\cite{dell2014automated,hofmann2014evolutionary,ballerini2008interaction,cavagna2010scalefree}. 
Vision is the only scalable, non-invasive way to recover such trajectories, while 2D analysis is insufficient for mechanistic modelling of spatial organisation and interactions~\cite{perezescudero2014idtracker,romeroferrero2019idtrackerai,herbertread2011inferring,calovi2015collective,perezescudero2013collective}.

Recovering dense \textit{3D trajectories} from video therefore remains a central bottleneck. 
A practical system must solve two coupled problems: \emph{spatial association}, which fuses multi-view 2D observations into 3D positions, and \emph{temporal association}, which links these positions into identity-consistent trajectories. 
For schooling fish, both are especially difficult because supervision is scarce and appearance cues are weak: cross-view identities, temporal annotations, and 3D ground truth are rarely available, while individuals look nearly identical and frequently occlude one another (Fig.~\ref{fig:schoolfish})~\cite{ispolatov2016collective,filella2018model,niwa1996newtonian}. 

\begin{wraptable}{r}{0.52\columnwidth}
    \vspace{-10pt}
    \centering
    \caption{Method families for dense 3D tracking. \textit{Temp./Spat.}: temporal/spatial association. The central columns indicate whether a family can operate without cross-view identity labels~(No XV-ID), temporal identity labels~(No T-ID), or 3D ground-truth trajectories~(No 3D\,GT). \textit{No Vis.}: no visual appearance features.}
    \label{tab:comparison}
    \vspace{2pt}
    \scriptsize
    \setlength{\tabcolsep}{1.6pt}
    \begin{tabular}{@{}lcc ccc c@{}}
        \toprule
        & & & \multicolumn{3}{c}{\shortstack{No labels /\\GT required}} & \\
        \cmidrule(lr){4-6}
        & Temp. & Spat. & \shortstack{No\\XV-ID} & \shortstack{No\\T-ID} & \shortstack{No 3D\\GT} & \shortstack{No\\Vis.} \\
        \midrule
        SVMOT & \checkmark & \ding{55} & \checkmark & \ding{55} & \checkmark & \ding{55} \\
        Supervised MVA & \ding{55} & \checkmark & \ding{55} & \checkmark & \checkmark & \ding{55} \\
        Self-supervised MVA & \ding{55} & \checkmark & \checkmark & \checkmark & \checkmark & \ding{55} \\
        MVMOT & \checkmark & \checkmark & \ding{55} & \ding{55} & \checkmark & \ding{55} \\
        \midrule
        \textbf{\name} & \checkmark & \checkmark & \checkmark & \checkmark & \checkmark & \checkmark \\
        \bottomrule
    \end{tabular}
    \vspace{-8pt}
\end{wraptable}
Existing vision approaches only partially address these requirements. 
\emph{Single-view multi-object trackers} (SVMOT)~\cite{memot2022,trackformer2022,cotracker2024,samba2024,motip2024} model temporal association within one camera, but cannot fuse observations across views. 
\emph{Multi-view association} (MVA) methods solve the complementary spatial problem by matching detections across synchronized cameras within each frame, but do not model temporal continuity; supervised variants such as MessyTable~\cite{messytable2020} and ViTP3DE~\cite{vitp3de2023} require cross-view identity labels, while self-supervised variants such as Self-MVA~\cite{selfmva2025} and MvMHAT~\cite{mvmhat2021,vo2020selfsupervised} still rely on visual features. 
\emph{Multi-view multi-object trackers} (MVMOT) such as ReST~\cite{rest2023}, MCTR~\cite{mctr2024}, LMGP~\cite{lmgp2022}, and DyGLIP~\cite{dyglip2021} jointly address spatial and temporal association, but require identity supervision for training. 
Thus, existing method families still depend on appearance features and/or identity supervision (Tab.~\ref{tab:comparison})---precisely the resources unavailable for dense fish schools.

In this paper, we propose \textbf{\name}, a self-supervised framework for the 3D association and tracking of visually homogeneous fish schools (Fig.~\ref{fig:densefish-method}). 
Our key insight is that, when individuals are nearly indistinguishable in appearance, reliable identity cues should come not from visual re-identification, but from the geometric and temporal consistency induced by calibrated multi-view observations. 
Accordingly, \name learns both cross-view correspondence and temporal identity consistency directly from multi-view geometry. 
Within each frame, it uses geometric cues to associate detections across cameras, triangulate candidate 3D hypotheses, and learn an embedding space in which detections of the same individual are pulled together while different individuals are separated. 
Across frames, it further learns temporal predictions that propagate identity embeddings through dense overlap and occlusions, reducing identity drift when instantaneous geometric matching becomes ambiguous. 
Triangulation and reprojection consistency provide the training signal throughout, turning multi-view consistency into supervision and eliminating the need for cross-view identity labels, temporal identity annotations, appearance-based re-identification, or ground-truth trajectories.
Here, self-supervised refers specifically to the 3D association and tracking stage.

To facilitate evaluation, we introduce \textbf{SynFish}, a realistic Unreal Engine~5 synthetic benchmark with dense ground-truth 3D trajectories for visually homogeneous fish schools.
We further evaluate on the real-world 3D-ZeF20 zebrafish benchmark~\cite{pedersen2020zef}.
Under a unified protocol that converts representative baselines into end-to-end 3D tracking pipelines, \name outperforms all methods on both datasets, achieving higher 3D tracking accuracy and fewer identity switches.

In summary, our main contributions are:
\begin{itemize}[leftmargin=*, itemsep=2pt, topsep=2pt, parsep=0pt]
\vspace{-2mm}
    \item A self-supervised multi-camera 3D tracking framework \textbf{\name} that jointly learns cross-view association and temporal identity consistency from calibrated multi-view geometry, requiring no cross-view identity labels, temporal identity annotations, appearance-based re-identification, or ground-truth 3D trajectories.

    \item \textbf{SynFish}, a realistic synthetic benchmark rendered in Unreal Engine~5, providing dense ground-truth 3D trajectories for visually homogeneous fish schools across varying group sizes, occlusion levels, and camera configurations.
\end{itemize}
\section{\textbf{Related Work}}

Dense multi-camera 3D tracking requires both \emph{spatial association} across views and \emph{temporal association} across frames. Existing methods address only part of this problem or rely on supervision and appearance cues, as summarized in Tab.~\ref{tab:comparison}.

\noindent\textbf{Multi-view association from 2D detections.}
Given per-camera detections, classical multi-view association matches observations across cameras and triangulates them into 3D~\cite{hartley1997triangulation}. These methods typically solve bipartite matching with Hungarian~\cite{kuhn1955hungarian} or greedy algorithms using epipolar and reprojection costs~\cite{hartley2003multiview}, but become fragile in dense scenes with occlusions, missed detections, and localization noise. Learned MVA methods replace hand-crafted costs with affinity models: supervised methods such as MessyTable~\cite{messytable2020} and ViTP3DE~\cite{vitp3de2023} require cross-view identity labels, while self-supervised methods such as Self-MVA~\cite{selfmva2025}, MvMHAT~\cite{mvmhat2021}, and Vo~et~al.~\cite{vo2020selfsupervised} reduce annotation needs. However, MVA methods only solve \emph{spatial} association within each frame and still rely on visual features, leaving temporal identity consistency to downstream heuristics; they do not learn a full 3D tracking pipeline over time.

\noindent\textbf{Single-view tracking followed by 3D lifting.}
Single-view multi-object trackers (SVMOT), including MeMOT~\cite{memot2022}, TrackFormer~\cite{trackformer2022}, CoTracker~\cite{cotracker2024}, SambaMOTR~\cite{samba2024}, and MOTIP~\cite{motip2024}, model temporal association within one camera using appearance, motion, or long-range tracklet cues. To obtain 3D trajectories, their per-view tracks must still be associated across cameras. For visually interchangeable animals, this produces fragmented 2D tracklets and reintroduces the same cross-view ambiguity faced by MVA methods.

\noindent\textbf{Joint multi-view multi-object tracking.}
MVMOT methods such as TRACTA~\cite{tracta2020}, DyGLIP~\cite{dyglip2021}, LMGP~\cite{lmgp2022}, ReST~\cite{rest2023}, and MCTR~\cite{mctr2024} jointly model spatial and temporal association using cross-camera re-identification with graph or transformer architectures. They are closest to full multi-camera tracking, but typically require identity supervision and discriminative appearance features. ADA-Track++~\cite{adatrack2024} further performs end-to-end multi-camera 3D tracking, but assumes 3D object detections and targets autonomous-driving scenes with visually distinct objects.

In contrast to these methods, \name targets the underexplored intersection of geometry-only identity association and self-supervised temporal 3D tracking for visually homogeneous animals by jointly learning cross-view association and temporal identity consistency from calibrated multi-view geometry, without identity labels, appearance-based re-identification, or ground-truth 3D trajectories.
\section{\textbf{Method}}\label{sec:method}

\name addresses dense multi-camera 3D multi-object tracking of schooling fish from synchronized videos captured by $C$ calibrated cameras. 
At each time step $t$, each camera provides 2D instance detections, each with a bounding box, an instance centroid $p=(u,v)$, and a detector confidence score $s\in[0,1]$; camera intrinsics $K_c$ and extrinsics $(R_c,t_c)$ are assumed known. 
\name is agnostic to the upstream 2D perception model: it only requires per-frame, per-camera 2D detections, which can come from a detector--segmenter pipeline such as YOLO26+SAM2~\cite{yolo26,sam2} or a promptable segmentation model such as SAM~3~\cite{sam3}.
It does not require temporal tracking output, clean detections, cross-view identity labels, or temporal identity labels from the 2D perception stage.
In our implementation, each per-camera, per-frame 2D instance is represented by a centroid, which is derived from an instance segmentation mask.
Overall, the method only assumes per-camera, per-frame 2D detections and calibrated cameras as input.

\begin{figure}[tbp]
    \vspace{-6mm}
    \centering
    \includegraphics[width=\columnwidth]{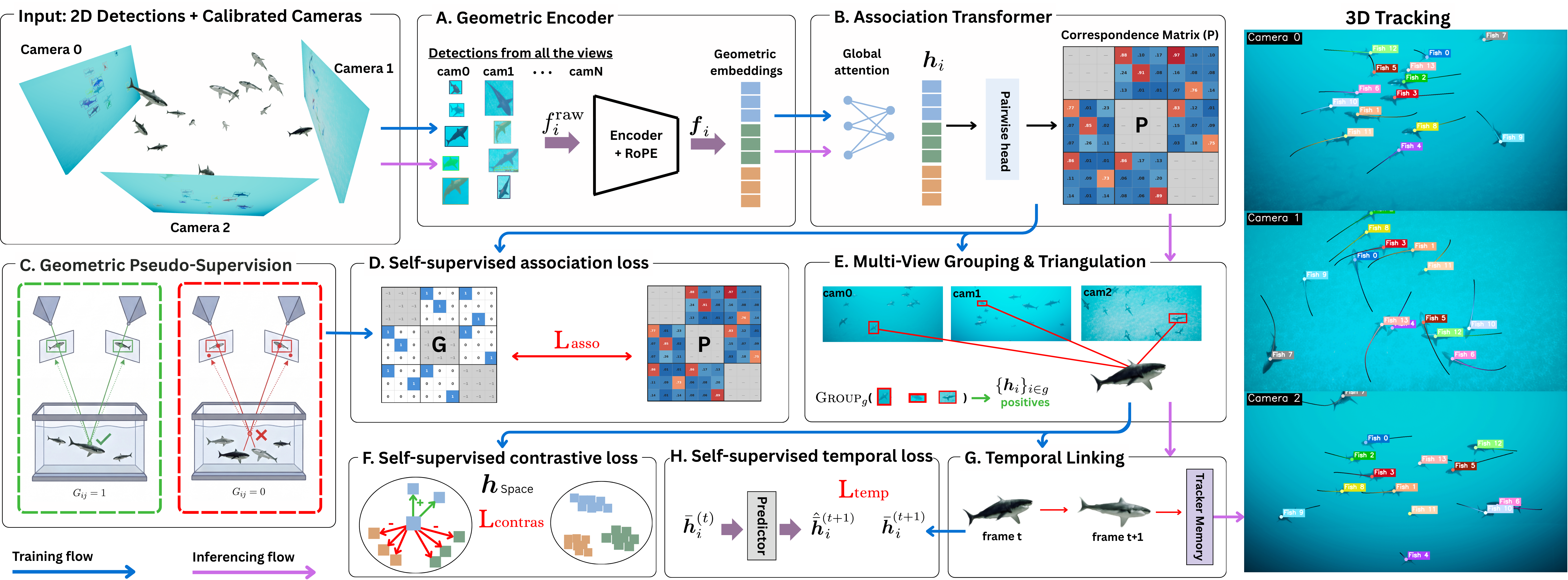}
    \caption{Overview of the \name framework. Synchronised multi-view video is processed by (A)~a \emph{Geometric Encoder} and (B)~an \emph{Association Transformer} that predicts cross-view pairwise scores, supervised by (C)~\emph{Geometric Pseudo-Supervision} from triangulation consistency through (D)~a \emph{Self-Supervised Association Loss}. Detections are fused via (E)~\emph{Multi-View Grouping \& Validation} to produce 3D hypotheses, whose embeddings are separated by (F)~a \emph{Self-Supervised Contrastive Loss}. Across frames, (G)~\emph{Temporal Linking} matches identities and (H)~a \emph{Self-Supervised Temporal Loss} enforces cross-frame consistency, enabling stable 3D trajectory recovery at inference.}
    \label{fig:densefish-method}
    \vspace{-5mm}
\end{figure}

As illustrated in Fig.~\ref{fig:densefish-method}, the 3D association and tracking stage of \name reconstructs trajectories self-supervisedly by learning three coupled components: cross-view association within each frame, embedding-space identity separation across views, and temporal identity consistency across frames. 
The complete training procedure is given in Algorithm~\ref{alg:densefish} (Appendix~\ref{app:algorithm}). 
Crucially, \name requires no cross-view identity annotations, no temporal identity annotations, and no ground-truth 3D trajectories.

\subsection{\textbf{Geometric encoder}}\label{sec:encoder}
Each raw detection feature, comprising the 2D centroid, bounding-box size, camera index, confidence score, and a back-projected world ray, is encoded by a lightweight MLP into a compact embedding that captures multi-view geometric constraints while remaining entirely agnostic to visual appearance.

Concretely, for a detection with mask-derived centroid $p=(u,v)$, bounding box size $(w,h)$, camera index $c$, and detector confidence $s\in[0,1]$, we construct a raw feature vector:
 $f_i^{\mathrm{raw}} = [\tilde{u}_i, \tilde{v}_i, \tilde{w}_i, \tilde{h}_i, \tilde{c}_i, s_i, r_{w,x}, r_{w,y}, r_{w,z}]$,
where centroid coordinates are divided by image width and height respectively, bounding-box dimensions are similarly normalised by the corresponding image dimension, and the camera index is divided by $(C{-}1)$, mapping all 2D quantities to approximately $[0,1]$.
The components $(r_{w,x}, r_{w,y}, r_{w,z})$ denote the unit world-ray direction obtained by back-projecting the centroid through the camera intrinsics and rotating into the world frame.
The geometric encoder $\phi$, a lightweight two-layer MLP
with ReLU~\cite{nair2010relu} and LayerNorm~\cite{ba2016layernorm} layers, maps this raw feature into a $d{=}128$-dimensional embedding:
 $\boldsymbol{f}_i = \phi(f_i^{\mathrm{raw}}) \in \mathbb{R}^d$.
This representation encodes geometric constraints while remaining agnostic to appearance.

\subsection{\textbf{Association transformer}}\label{sec:transformer}
For each frame, the encoder outputs $\{\boldsymbol{f}_i\}$ are processed jointly by an association transformer $\psi$, producing context-enriched embeddings $\{\boldsymbol{h}_i\}$.
We adopt a \emph{global} self-attention design over all $N$ detections from all $C$ cameras in a frame, similar in spirit to prior geometric token processing in VGGT~\cite{vggt2025}, rather than applying cross-attention between camera pairs.
Detections from all cameras are processed jointly so that correspondence decisions can exploit global cross-view context within each frame. This global self-attention design captures all-to-all relationships in one pass, supports transitive reasoning across cameras, and avoids separate pair-specific computations. We inject 2D spatial information using rotary positional embeddings (RoPE)~\cite{rope2024} derived from normalised centroid coordinates, which encode relative spatial structure without fixed resolution assumptions.
This scene-level context is especially important in dense scenes, where multiple cross-view candidates may satisfy similar pairwise geometric costs. Rather than deciding each match independently, the transformer conditions each detection on the full set of observations in the frame before multi-view grouping, helping disambiguate geometrically plausible but incorrect matches.

A pairwise association head estimates whether detections from different cameras correspond to the same individual. For each ordered pair of detections $(i,j)$ from different cameras, the head outputs a probability $P_{ij}\in(0,1)$ indicating whether the two detections correspond to the same fish.
Pairs from the same camera are ignored by construction.

Concretely, let $\boldsymbol{x}_i = \psi_{\mathrm{proj}}(\boldsymbol{f}_i)$ denote the projected token for detection $i$, and let $\boldsymbol{p}_i=(c_{y,i},\,c_{x,i})\in[0,1]^2$ be its normalised centroid. In the implementation, RoPE is applied to the projected token \emph{before} the transformer encoder layers:
 $\boldsymbol{x}_i' = R(\boldsymbol{p}_i)\,\boldsymbol{x}_i$,
where the vertical and horizontal centroid axes are encoded independently in separate halves of the embedding dimension. The rotated tokens $\{\boldsymbol{x}_i'\}$ are then processed by the standard transformer self-attention stack to produce contextualised embeddings $\{\boldsymbol{h}_i\}$. Because the rotary transform injects position information in a relative, parameter-free manner, the model can distinguish spatially nearby detections from distant ones without a fixed learned positional embedding table, which is particularly important for disambiguating nearby fish in dense scenes.

\subsection{\textbf{Geometric pseudo-supervision}}\label{sec:pseudogt}
Because cross-view identity annotations are unavailable, \name generates pseudo-labels through multi-view geometric consistency.
The contribution here is not a new triangulation primitive, but using calibrated multi-view geometry as supervision for cross-view association without identity labels.
For each pair of detections from different cameras within the same timestamp 
, we compute a tentative 3D point by triangulation~\cite{hartley2003multiview} and reproject it back into the image planes.
If two detections correspond to the same fish, triangulation is geometrically consistent and reprojections align with the original observations.
Based on this geometric consistency criterion, we construct a binary pseudo ground-truth association matrix $G \in \{0,1\}^{N\times N}$.
For a cross-camera detection pair $(i,j)$, we assign $G_{ij}=1$ if the triangulated point reprojects inside both bounding boxes, and $G_{ij}=0$ otherwise.
Pairs from the same camera and self-pairs are excluded from supervision and masked out during training.

To account for geometric noise, we additionally compute a confidence score for positive pairs based on reprojection error
$C_{ij} = \frac{1}{1+\mathrm{err}_{ij}}$, where $\mathrm{err}_{ij}$ denotes the average reprojection error \emph{in pixels} across the two views (pseudo-GT computation operates in raw pixel coordinates; the normalisation described in Section~\ref{sec:encoder} applies only inside the feature encoder).
This confidence modulates the strength of supervision without changing the binary nature of the pseudo labels.

\subsection{\textbf{Self-supervised association loss}}\label{sec:loss_asso}
The association loss supervises the pairwise correspondence head using the geometry-derived binary pseudo labels.
For each frame, we consider ordered detection pairs $(i,j)$ such that $i\neq j$ and $c_i\neq c_j$.
Pairs excluded during pseudo-GT construction are masked out and do not contribute to the loss.

Let $P_{ij}\in(0,1)$ denote the predicted correspondence probability and $G_{ij}\in\{0,1\}$ the pseudo label.
We apply a confidence-weighted binary cross-entropy loss in which positive and negative terms are normalised independently:
 $\mathcal{L}_{\mathrm{asso}} = -\frac{\sum_{G_{ij}=1} C_{ij}\,\log P_{ij}}{\sum_{G_{ij}=1} C_{ij}} - \frac{\sum_{G_{ij}=0} \log(1{-}P_{ij})}{|\{G_{ij}=0\}|}$,
where $C_{ij}$ is the pseudo-GT confidence (Section~\ref{sec:pseudogt}).
Positive pairs are weighted by their confidence and averaged by the total confidence mass, while negative pairs receive uniform weight and are averaged by count.
This independent normalisation prevents negative-pair domination when positive pairs are scarce, ensuring balanced gradient contributions regardless of the positive-to-negative ratio.

This loss is applied independently to frames $t$ and $t{+}1$ during training.
It encourages high correspondence probability for geometrically consistent cross-view pairs and low probability otherwise, providing a self-supervised signal for learning multi-view association without identity labels or ground-truth 3D trajectories.

\subsection{\textbf{Multi-view grouping \& triangulation}}\label{sec:grouping}
Predicted correspondence probabilities are converted into multi-view fish hypotheses through a compact grouping pipeline. For cross-camera pairs, we combine learned correspondence and embedding agreement with
 $A_{ij} = w_p\,P_{ij} + w_f\,\frac{\cos(\boldsymbol{h}_i, \boldsymbol{h}_j)+1}{2}$,
then apply Hungarian matching within each camera pair, merge accepted matches transitively across cameras, and triangulate each candidate group. Groups are kept only when the triangulated 3D point reprojects consistently into the supporting detections. For each valid group, we compute the group embedding
 $\bar{\boldsymbol{h}}_g = \ell_2\text{-norm}\!\bigl(\tfrac{1}{|g|}\sum_{i\in g}\boldsymbol{h}_i\bigr)$,
During training, grouping is not a supervision target. Instead, it produces two outputs used downstream: the individual member embeddings $\{\boldsymbol{h}_i\}_{i\in g}$, which define positive pairs for $\mathcal{L}_{\mathrm{ctr}}$ (Section~\ref{sec:loss_ctr}), and the group embedding $\bar{\boldsymbol{h}}_g$, which serves as the per-fish identity token for temporal linking (Section~\ref{sec:loss_temp}). Full grouping, thresholding, and geometric validation details are provided in Appendix~\ref{app:extended_grouping}.

\subsection{\textbf{Self-supervised contrastive loss}}\label{sec:loss_ctr}
Within each frame, detections from the same multi-view group are treated as positives and all others as negatives. We apply an InfoNCE-style contrastive loss~\cite{oord2018infonce} to the normalised transformer embeddings $\boldsymbol{z}_i = \boldsymbol{h}_i / \|\boldsymbol{h}_i\|$, with similarity $S_{ij} = \boldsymbol{z}_i^\top \boldsymbol{z}_j / \tau$.
For each anchor detection $i$ with positive set $\mathcal{P}_i$, the loss is:
 $\mathcal{L}_{\mathrm{ctr}}^{(i)} = -\log \frac{\sum_{j\in\mathcal{P}_i}\exp(S_{ij})}{\sum_{\substack{k=1\\k\neq i}}^{N}\exp(S_{ik})}$,
and the frame-level loss averages over anchors with at least one positive. This objective pulls same-fish detections together across views and pushes different identities apart, improving grouping in dense scenes (Fig.~\ref{fig:contrastive-impact}). Full positive-mining and implementation details are provided in Appendix~\ref{app:extended_contrastive}.

\subsection{\textbf{Temporal linking and self-supervised temporal loss}}\label{sec:loss_temp}
To encourage identity stability over time, \name links triangulated group hypotheses between consecutive frames and trains a temporal predictor on the matched group embeddings. Given groups from frames $t$ and $t{+}1$ (each with a 3D position $\mathbf{X}$ and a group embedding $\bar{\boldsymbol{h}}$), we construct the matching cost
 $\mathrm{cost}(i,j) = \alpha\,\frac{\|\mathbf{X}_j^{(t+1)} - \mathbf{X}_i^{(t)}\|}{d_{\max}} + \beta\,\big(1 - \cos(\bar{\boldsymbol{h}}_i^{(t)}, \bar{\boldsymbol{h}}_j^{(t+1)})\big)$,
where $\alpha{=}0.6$ and $\beta{=}0.4$ balance 3D proximity and embedding similarity. Hungarian assignment~\cite{kuhn1955hungarian} is applied with distance and similarity gates, and each accepted match receives a weight $w_{ij} = \min(\mathrm{conf}_i, \mathrm{conf}_j)$.
For each matched pair $(i,j)$, the predictor takes the observed group embedding $\bar{\boldsymbol{h}}_i^{(t)}$ and predicts $\hat{\bar{\boldsymbol{h}}}_j^{(t+1)} = \rho(\bar{\boldsymbol{h}}_i^{(t)})$. The temporal loss is a weight-normalized mean squared error:
 $\mathcal{L}_{\mathrm{temp}} = \frac{\sum_{(i,j)} w_{ij}\,\|\hat{\bar{\boldsymbol{h}}}_j - \bar{\boldsymbol{h}}_j^{(t+1)}\|^2}{\sum_{(i,j)} w_{ij}}$,
which encourages identity-consistent embeddings across time and improves linking under occlusion (Fig.~\ref{fig:temporal-impact}). Full fallback matching, predictor, and training details are provided in Appendix~\ref{app:extended_temporal}.

\subsection{\textbf{Training and Inference}}
\label{sec:inference}
The total training loss is a weighted combination of the three complementary objectives:
$$
\mathcal{L} = \mathcal{L}_{\mathrm{asso}} + \lambda_{\mathrm{ctr}}\,\mathcal{L}_{\mathrm{ctr}} + \lambda_{\mathrm{temp}}\,\mathcal{L}_{\mathrm{temp}}.
$$
Training begins with the association loss alone and gradually introduces the contrastive and temporal terms to stabilize optimization.
Each component targets a distinct failure mode: $\mathcal{L}_{\mathrm{asso}}$ resolves cross-view ambiguity, $\mathcal{L}_{\mathrm{ctr}}$ reduces within-frame identity confusion, and $\mathcal{L}_{\mathrm{temp}}$ prevents temporal identity drift.
No component requires identity annotations or ground-truth 3D data; all supervision is derived from multi-view geometric consistency.

At inference, \name reconstructs per-frame 3D hypotheses using the learned association scores $P_{ij}$ and embeddings, then links them over time with the same distance--similarity cost and the learned predictor $\rho$. Missing tracks are bridged by rolling embeddings forward with $\rho$, and non-overlapping fragments are stitched using embedding similarity and 3D proximity. Optional interpolation and short-track removal are applied only as post-processing. Thus, inference is a direct continuation of training and requires no identity labels, appearance re-identification, or privileged scene information. Full tracker and stitching details are provided in Appendix~\ref{app:extended_inference}.

\begin{figure}[tbp]
    \vspace{-6mm}
    \centering
    \begin{minipage}[t]{0.48\columnwidth}
        \centering
        \vspace{0pt}
        \includegraphics[width=\linewidth,height=0.24\textheight,keepaspectratio]{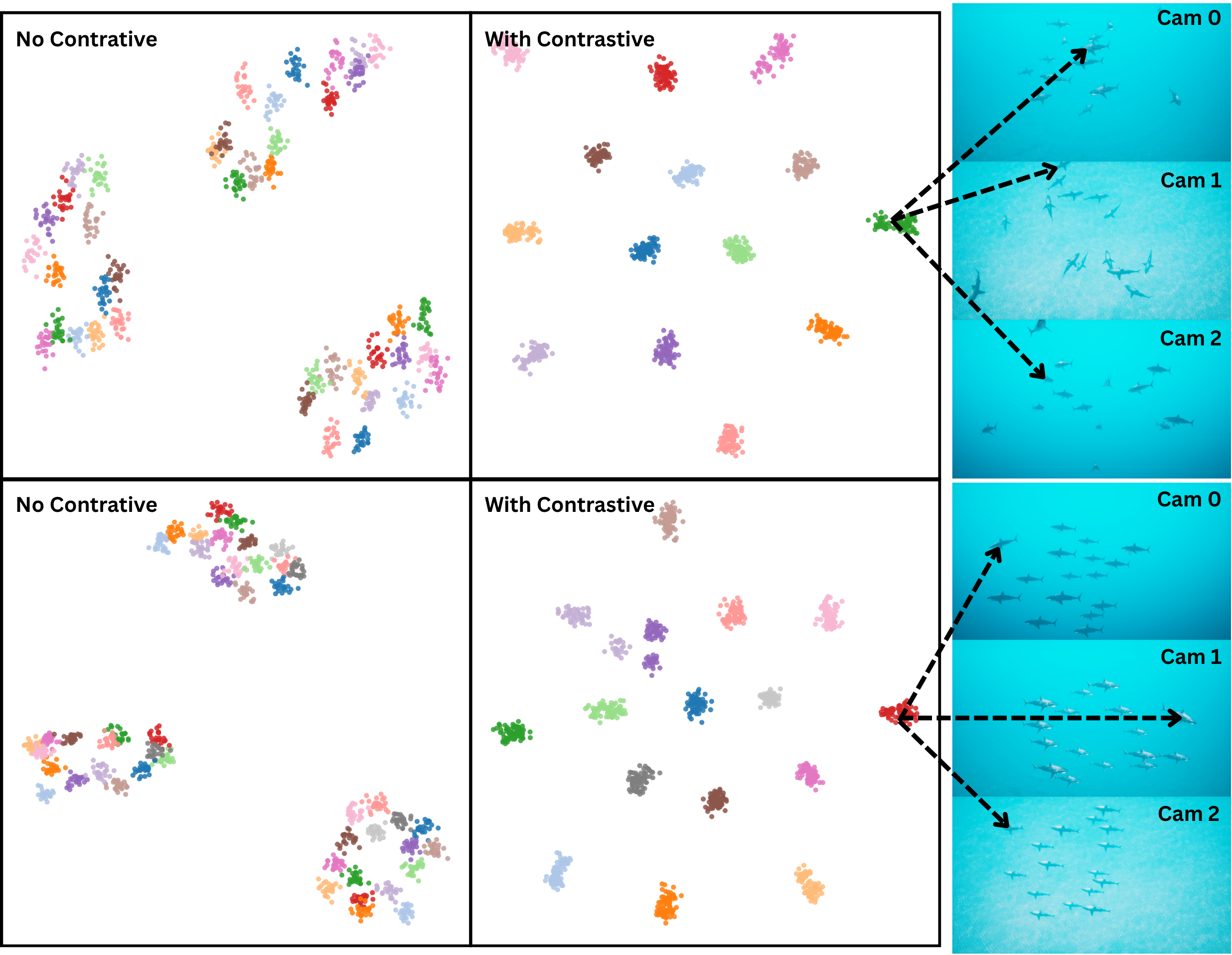}
        \caption{Effect of the contrastive objective on cross-view embeddings. \textbf{Left:} Without contrastive training, same-fish detections overlap other identities. \textbf{Right:} With contrastive training, they form compact, well-separated clusters.}
        \label{fig:contrastive-impact}
    \end{minipage}
    \hfill
    \begin{minipage}[t]{0.48\columnwidth}
        \centering
        \vspace{0pt}
        \includegraphics[width=\linewidth,height=0.24\textheight,keepaspectratio]{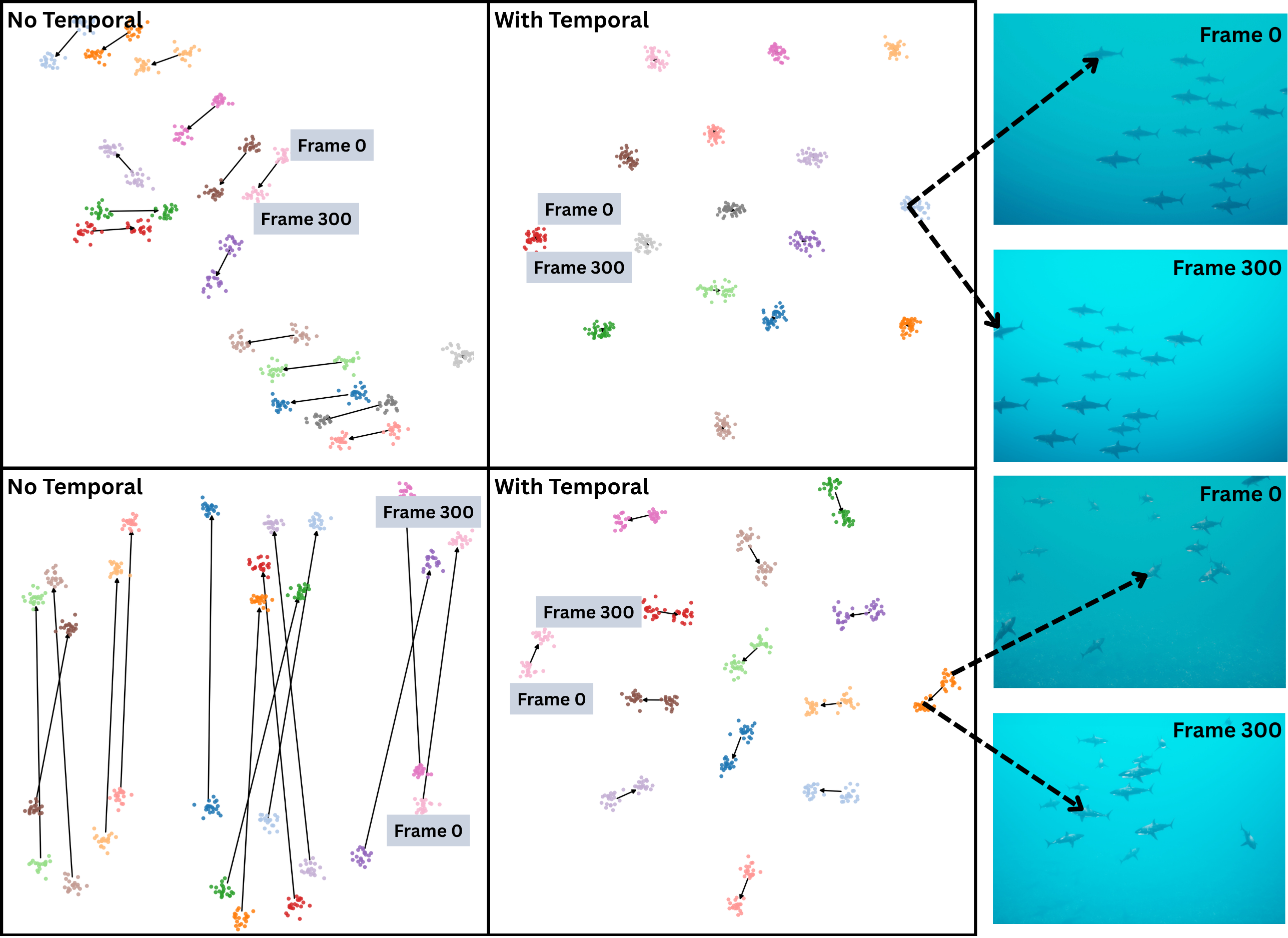}
        \caption{Effect of the temporal prediction objective on cross-frame identity consistency. \textbf{Left:} Without temporal training, same-fish embeddings drift across frames and cause identity switches. \textbf{Right:} With temporal training, they stay close across frames for stable linking.}
        \label{fig:temporal-impact}
    \end{minipage}
    \vspace{-5mm}
\end{figure}


\section{\textbf{Experimental Results}}\label{sec:eval}

\subsection{\textbf{Datasets and difficulty characterisation}}\label{sec:eval_dataset}

We evaluate \name on the synthetic \textbf{SynFish} benchmark and the real-world zebrafish dataset 3D-ZeF~\cite{pedersen2020zef}. All datasets provide calibrated cameras; SynFish additionally provides ground-truth 3D trajectories for quantitative evaluation.

\noindent\textbf{SynFish dataset.}
SynFish contains 14 scenes with 4--16 visually near-indistinguishable fish, rendered at $1920{\times}1080$ with $C{=}3$ synchronized cameras and 7 camera-rig configurations. We train on 8 scenes and evaluate on 6 held-out scenes, including one unseen rig (Appendix~\ref{app:cam_ue5}).

\noindent\textbf{Real-world zebrafish (3D-ZeF).}
3D-ZeF~\cite{pedersen2020zef} provides calibrated two-camera real-world sequences at $2704{\times}1520$ resolution with 2 or 5 fish. We use zebra1 and zebra3 for training and zebra2 and zebra4 for testing.

To characterise difficulty beyond fish count, we report an occlusion score (OccScore), defined as the average maximum pairwise 2D bounding-box IoU over frames and views, together with \%Overlap, the percentage of frame-view instances containing any overlap (Appendix Table~\ref{tab:dataset_summary}). Camera calibration parameters are listed in Appendix~\ref{app:camera_params}. Please refer to Appendix~\ref{app:extended_dataset}--\ref{app:extended_impl} for more details.

\subsection{\textbf{Baselines, Protocol, and Metrics}}\label{sec:eval_baselines}

No prior method solves this setting end-to-end without identity supervision, so we compare against 13 complete 3D tracking pipelines spanning four families. 
Detection-based baselines combine SAM~3~\cite{sam3} or YOLO26+SAM2~\cite{yolo26,sam2} detections with geometric MVA (Hungarian~\cite{kuhn1955hungarian} or greedy) or learned MVA (\emph{Self-MVA}~\cite{selfmva2025}, \emph{ASNet}~\cite{messytable2020}), followed by triangulation and nearest-neighbour 3D linking; \textbf{3D-SORT} instead adds constant-velocity Kalman tracking~\cite{bewley2016simple} after SAM3 + Hungarian (Appendix~\ref{app:sort_baseline}).
We also evaluate single-view MOT methods, \emph{SambaMOTR}~\cite{samba2024} and \emph{MOTIP}~\cite{motip2024}, lifted to 3D via first-frame cross-view association, and multi-camera MOT methods, \emph{ReST}~\cite{rest2023} and \emph{MCTR}~\cite{mctr2024}, whose cross-view-consistent 2D tracks are triangulated directly.

For fairness, all external components are finetuned or retrained on the SynFish and 3D-ZeF training splits with default configurations, while \name is trained self-supervisedly on the same splits using SAM~3 or YOLO26+SAM2 detections. 
All pipelines use identical detections, camera calibrations, and post-processing, including short-gap interpolation between already-associated observations and removal of short-lived ghost tracks; these steps use no privileged information and introduce no new identities. 
We evaluate reconstructed trajectories using standard 3D MOT metrics: MOTA, mostly-tracked (MT\%), mostly-lost (ML\%), identity switches (IDSw), fragmentations (Frag), and mean time between failures (MTBF). 
Predicted 3D points are matched to ground truth under a fixed 3D distance threshold at each frame, from which trajectory-level metrics are computed.

\subsection{\textbf{Results on SynFish test set}}\label{sec:eval_ue5}

\begin{figure*}[t]
    \vspace{-6mm}
\centering
\begin{minipage}[t]{0.56\textwidth}
\vspace{0pt}
\centering
\captionof{table}{3D tracking results on the SynFish test set, averaged over 6 held-out scenes (4--16 fish). Best result in each column among methods with MOTA${>}0$ is \textbf{bolded}; per-scene breakdown is in Appendix~Table~\ref{tab:app_ue5_perscene}.}
\label{tab:ue5_test_avg}
\resizebox{\linewidth}{!}{
\begin{tabular}{l rrrrrr}
\hline
\textbf{Method} & \textbf{MOTA\,$\uparrow$} & \textbf{MT\%\,$\uparrow$} & \textbf{ML\%\,$\downarrow$} & \textbf{IDSw\,$\downarrow$} & \textbf{Frag\,$\downarrow$} & \textbf{MTBF\,$\uparrow$} \\
\hline
\multicolumn{7}{l}{\emph{Detection + geometric MVA + 3D nearest-neighbour linking}} \\
SAM3 + Hungarian                      & 83.5\% & 71.9\% & 0.0\%  &  7.5 & 12.8 & 183.8 \\
SAM3 + Greedy                         & 83.5\% & 71.9\% & 0.0\%  &  7.5 & 12.8 & 183.8 \\
SAM3 + 3D-SORT                        & 87.7\% & 77.1\% & 0.0\%  &  4.0 &  9.0 & 209.4 \\
YOLO26+SAM2 + Hungarian               & 82.2\% & 69.8\% & 0.0\%  &  8.7 & 12.2 & 183.1 \\
YOLO26+SAM2 + Greedy                  & 82.2\% & 69.8\% & 0.0\%  &  8.7 & 12.2 & 183.1 \\
\hline
\multicolumn{7}{l}{\emph{Detection + learned MVA + 3D nearest-neighbour linking}} \\
Self-MVA~\cite{selfmva2025}           & $-$64.6\% &  0.0\% & 96.9\% &  1.7 &  2.2 &  15.9 \\
ASNet~\cite{messytable2020}           &  33.3\% & 49.6\% & 11.5\% & 20.8 & 29.0 &  60.7 \\
\hline
\multicolumn{7}{l}{\emph{Single-view 2D MOT $\rightarrow$ MVA (first frame) $\rightarrow$ 3D}} \\
SambaMOTR~\cite{samba2024}            & $-$10.5\% & 11.7\% & 73.8\% &  0.0 &  7.8 &  51.8 \\
MOTIP~\cite{motip2024}                &  14.7\% & 21.4\% & 42.9\% &  0.0 & 25.7 &  41.2 \\
\hline
\multicolumn{7}{l}{\emph{Multi-camera MOT $\rightarrow$ triangulate}} \\
ReST~\cite{rest2023}                  &  61.3\% & 74.8\% &  0.0\% & 173.7 & 78.5 &  17.5 \\
MCTR~\cite{mctr2024}                  &  83.7\% & 82.7\% &  0.0\% &  1.7 & 14.2 & 160.6 \\
\hline
\multicolumn{7}{l}{\emph{Ours (self-supervised 3D tracking)}} \\
\name (SAM3)                      &  95.8\% & \textbf{95.8\%} & \textbf{0.0\%} & \textbf{0.0} &  6.5 & 266.3 \\
\name (YOLO26+SAM2)              & \textbf{96.6\%} & \textbf{95.8\%} & \textbf{0.0\%} &  0.2 & \textbf{3.7} & \textbf{296.5} \\
\hline
\end{tabular}
}
\end{minipage}
\hfill
\begin{minipage}[t]{0.40\textwidth}
\vspace{0pt}
\centering
\includegraphics[width=\linewidth]{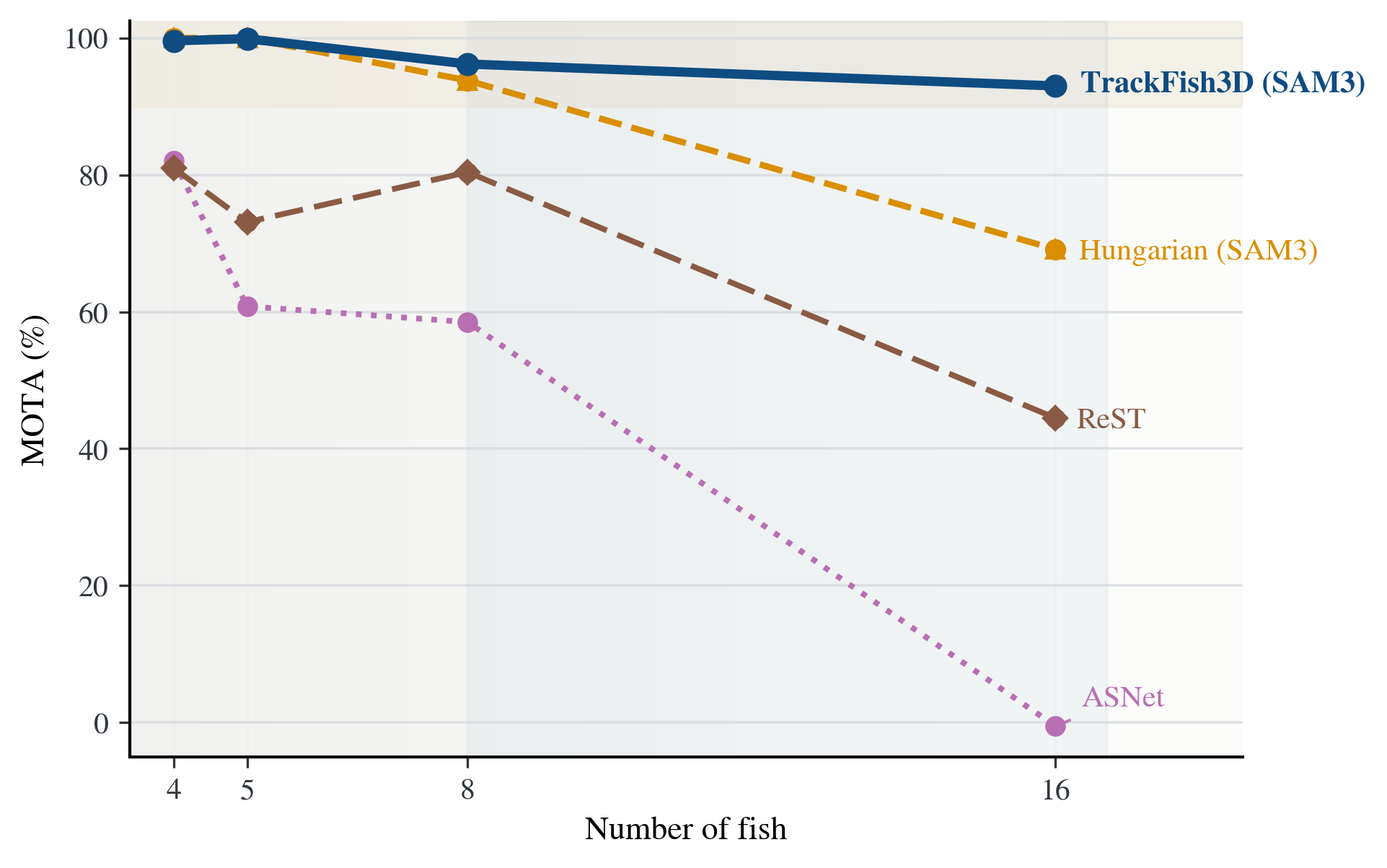}
\captionof{figure}{MOTA vs.\ fish count on SynFish test scenes. \name degrades gracefully with density, whereas baselines drop sharply beyond 8 fish.}
\label{fig:mota_vs_fish}
\end{minipage}
\end{figure*}

Table~\ref{tab:ue5_test_avg} presents 3D tracking results averaged over the 6 held-out SynFish test scenes (4--16 fish).
Per-scene breakdowns are provided in Appendix~Table~\ref{tab:app_ue5_perscene}.

Across six held-out SynFish scenes, \name achieves the best overall performance, reaching 96.6\% MOTA with YOLO26+SAM2 inputs and 95.8\% with SAM3, while also obtaining the highest MT\% (95.8\%) and strongest identity stability (IDSw~0.0--0.2, MTBF~266--297 frames). 
It consistently outperforms all baselines, including supervised methods retrained on the same data. 
The strongest baseline, 3D-SORT, reaches 87.7\% MOTA; \emph{MCTR} reaches 83.7\%, and direct geometric MVA pipelines achieve 82--84\%, while \name remains above 86\% on dense 16-fish scenes.
In contrast, appearance-driven and single-view trackers fail to transfer reliably to visually homogeneous fish. 
Fig.~\ref{fig:mota_vs_fish} further shows that \name degrades gracefully with fish density, whereas baselines drop sharply beyond 8 fish.

\begin{figure}[t]
    \vspace{-2mm}
    \centering
    \includegraphics[width=1\columnwidth]{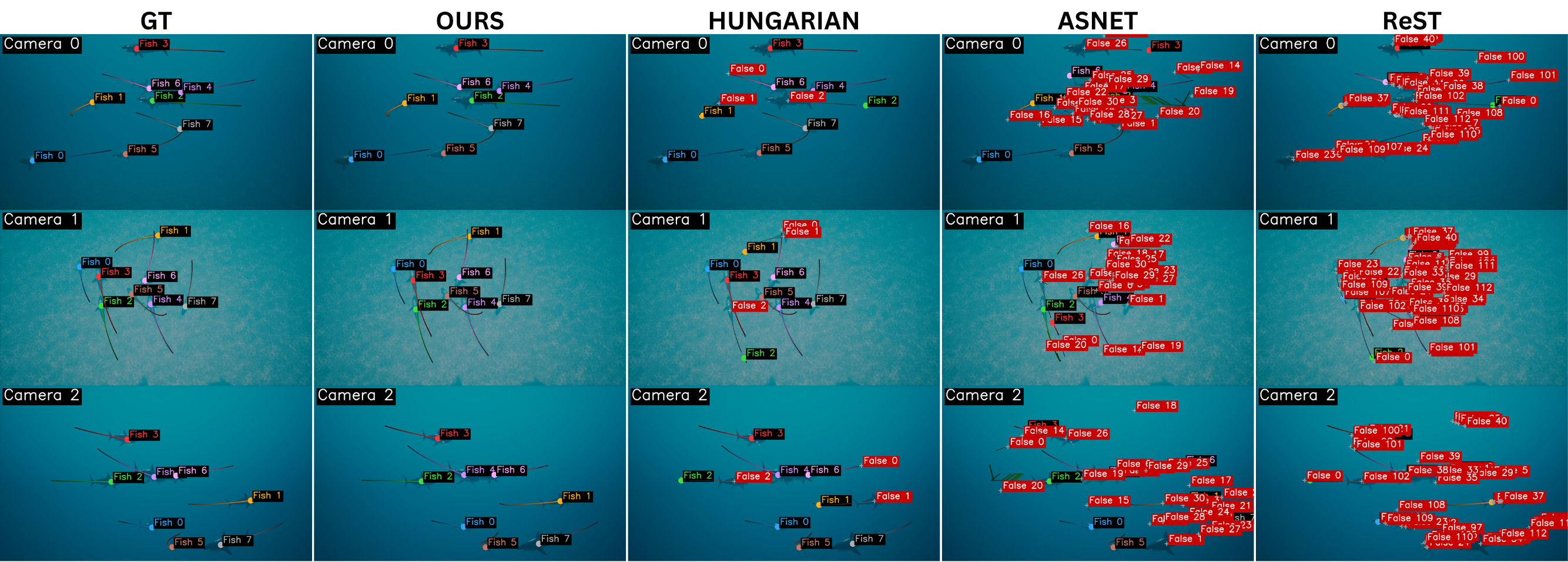}
    \caption{Qualitative 3D trajectory comparison on SynFish test scenes. Columns: ground truth, \name, and representative baselines. Colour encodes fish identity; trajectory breaks indicate identity switches or fragmentations.}
    \label{fig:qualitative_ue5}
    \vspace{0mm}
\end{figure}

\subsection{\textbf{Results on 3D-ZeF test set}}\label{sec:eval_real}

\begin{table}[t]
    \vspace{0mm}
\centering
\caption{3D tracking results on the 3D-ZeF test set, averaged over 2 sequences (zebra2: 5 fish, zebra4: 5 fish).
Best result in each column (among methods with MOTA${>}0$) is \textbf{bolded}.
}
    \vspace{1mm}
\label{tab:zebra_test_avg}
\resizebox{0.9\columnwidth}{!}{
\begin{tabular}{l rrrrrr}
\hline
\textbf{Method} & \textbf{MOTA\,$\uparrow$} & \textbf{MT\%\,$\uparrow$} & \textbf{ML\%\,$\downarrow$} & \textbf{IDSw\,$\downarrow$} & \textbf{Frag\,$\downarrow$} & \textbf{MTBF\,$\uparrow$} \\
\hline
Hungarian + nearest                   &  77.4\% & \textbf{90.0\%} &  \textbf{0.0\%} &   6.0 &  73.0 &  42.5 \\
Greedy + nearest                      &  76.9\% & \textbf{90.0\%} &  \textbf{0.0\%} &  10.0 &  76.0 &  38.7 \\
3D-SORT                               &  75.6\% & \textbf{90.0\%} &  \textbf{0.0\%} &  13.5 &  65.5 &  43.4 \\
Self-MVA~\cite{selfmva2025}           &  50.8\% &  60.0\% &  \textbf{0.0\%} &  21.5 &  73.5 &  48.5 \\
ASNet~\cite{messytable2020}           &   0.3\% &   0.0\% &  \textbf{0.0\%} &  72.5 & 175.0 &  10.6 \\
SambaMOTR~\cite{samba2024}            & $-$26.4\% &  0.0\% & 100.0\% &   0.5 &   1.0 &   6.5 \\
MOTIP~\cite{motip2024}                & $-$45.9\% &  0.0\% & 100.0\% &   1.5 &   6.5 &   2.7 \\
ReST~\cite{rest2023}                  &   0.0\% &   0.0\% & 100.0\% &   0.0 &   0.0 &   0.0 \\
MCTR~\cite{mctr2024}                  &   0.1\% &   0.0\% &  40.0\% &  21.0 &  57.5 &   8.8 \\
\hline
\name                             & \textbf{81.1\%} & \textbf{90.0\%} & \textbf{0.0\%} & \textbf{4.0} & \textbf{30.5} & \textbf{93.6} \\
\hline
\end{tabular}
}
    \vspace{-4mm}
\end{table}
Tab.~\ref{tab:zebra_test_avg} reports results on the real 3D-ZeF test set, where \name is trained self-supervisedly on zebra1/3 and evaluated on zebra2/4 using the same geometric objectives as on SynFish; per-scene results are in Appendix~Table~\ref{tab:app_zebra_perscene}. 
\name achieves 81.1\% MOTA, outperforming all baselines. 
The gap over geometric MVA baselines is modest in MOTA (77.4\% for Hungarian, 76.9\% for Greedy) because 3D-ZeF contains only 5 fish, making per-frame cross-view association relatively sparse.

The advantage is clearer in identity stability: \name reduces fragmentation from 73--76 to 30.5, improves MTBF from 38--43 to 93.6 frames, and lowers ID switches from 6--10 to 4.0. 
This shows the benefit of learned temporal linking under real occlusions, where per-frame geometric matching alone is insufficient. 
Appearance-based baselines remain ineffective after retraining, with most producing near-zero or negative MOTA and Self-MVA reaching only 50.8\%, again indicating that geometry is the more reliable cue for visually homogeneous animals.

\begin{figure}[t]
    \centering
    \includegraphics[width=0.8\columnwidth]{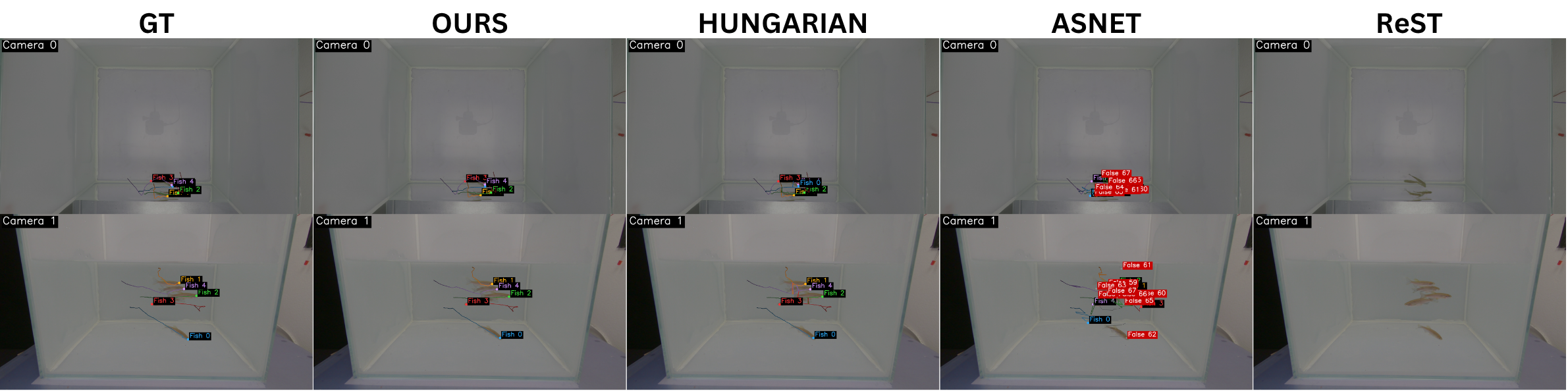}
    \caption{Qualitative 3D trajectory comparison on 3D-ZeF test sequences. \name maintains identity consistency through real-world occlusions that fragment or swap baseline trajectories. ReST produces no trajectories because its cross-view association fails to match detections correctly.}
    \label{fig:qualitative_zebra}
    \vspace{-6mm}
\end{figure}

\subsection{Cross-Species Generalization to Real-World Pigeons}
\label{sec:eval_pigeon}

On the real-world Pigeon10 subset from 3D-POP~\cite{naik2023pop}, \name achieves the best MOTA (91.9\%), recall (94.5\%), and fragmentation (4) among evaluated methods without bird identity labels. We emphasise that this subset mainly captures pigeons walking on the ground, with only limited flying in or out of view, rather than sustained 3D flocking. We therefore present it as preliminary evidence that the geometry-driven formulation can transfer beyond fish to another visually homogeneous animal setting, not as validation on fully 3D flying-bird tracking; evaluating on harder datasets with sustained bird flight is left for future work. Full setup and results are provided in Appendix~\ref{app:pigeon}.

\subsection{\textbf{Ablation Studies}}\label{sec:eval_ablation}
We conduct ablation studies on SynFish to isolate (i)~the contribution of each self-supervised loss component and (ii)~the impact of visual appearance features.
Results are reported in Tab.~\ref{tab:ablation}.
\begin{table}[H]
\vspace{-5mm}
\centering
\caption{Ablation studies on SynFish.
\emph{Top:} removing contrastive ($\mathcal{L}_{\mathrm{ctr}}$) and/or temporal ($\mathcal{L}_{\mathrm{temp}}$) losses.
\emph{Bottom:} concatenating DINOv3~\cite{dinov3} visual features with geometric features.}
\label{tab:ablation}
\vspace{1mm}
\resizebox{0.80\columnwidth}{!}{
\begin{tabular}{l rrrrrr}
\hline
\textbf{Variant} & \textbf{MOTA\,$\uparrow$} & \textbf{MT\%\,$\uparrow$} & \textbf{ML\%\,$\downarrow$} & \textbf{IDSw\,$\downarrow$} & \textbf{Frag\,$\downarrow$} & \textbf{MTBF\,$\uparrow$} \\
\hline
\multicolumn{7}{l}{\emph{Ablation 1: loss components}} \\
\name (full)                      & \textbf{97.7\%} & \textbf{98.3\%} & \textbf{0.0\%}  & \textbf{0.5} & \textbf{3.1} & \textbf{283.1} \\
\quad w/o $\mathcal{L}_{\mathrm{temp}}$            &  86.0\% & 85.2\% &  8.0\% &  2.0 &  5.2 & 243.1 \\
\quad w/o $\mathcal{L}_{\mathrm{ctr}}$             &  36.8\% & 23.8\% & 36.4\% &  3.0 &  8.0 &  99.1 \\
\quad w/o both                                     &  36.9\% & 25.6\% & 34.2\% &  3.8 &  9.8 &  88.8 \\
\hline
\multicolumn{7}{l}{\emph{Ablation 2: visual features}} \\
\name (geometry only)             & 97.7\% & 98.3\% & 0.0\% & \textbf{0.5} & \textbf{3.1} & \textbf{283.1} \\
\name + DINOv3                    & \textbf{97.9}\% & \textbf{99.2}\% & 0.0\% & 1.1 & 3.2 & 270.2 \\
\hline
\end{tabular}
}
\vspace{-4mm}
\end{table}

\noindent\textbf{Different Loss Components.}
Removing the contrastive loss ($\mathcal{L}_{\mathrm{ctr}}$) causes a significant drop from 97.7\% to 36.8\% MOTA, with ML\% rising from 0\% to 36.4\%.
Removing the temporal loss ($\mathcal{L}_{\mathrm{temp}}$) reduces MOTA from 97.7\% to 86.0\% and roughly quadruples identity switches (0.5$\rightarrow$2.0); spatial association still functions, but occlusion-bridging degrades noticeably, leading to higher fragmentation and lower MTBF.
Removing both auxiliary losses yields 36.9\% MOTA, nearly identical to removing $\mathcal{L}_{\mathrm{ctr}}$ alone, indicating that without view-consistent identity separation, temporal prediction has little to anchor to.

\noindent\textbf{Influence of Visual Features.}
Concatenating pretrained DINOv3~\cite{dinov3} visual features with the geometric features yields 97.9\% MOTA versus 97.7\% for geometry alone, a negligible difference.
MTBF actually decreases slightly (270.2 vs.\ 283.1), and IDSw doubles (1.1 vs.\ 0.5), indicating no practical gain from adding appearance cues. 

\noindent\textbf{Controlled Association Transformer Ablation.}
To isolate the role of scene-level self-attention, we replace the Association Transformer with an MLP-only encoder while retaining the same training data and all three losses. Results are reported in Tab.~\ref{tab:attention_ablation}.

\begin{table}[H]
\centering
\caption{Controlled Association Transformer ablation on SynFish.}
\label{tab:attention_ablation}
\small
\begin{tabular}{lrrrrr}
\hline
\textbf{Variant} & \textbf{MOTA\,$\uparrow$} & \textbf{MT\%\,$\uparrow$} & \textbf{IDSw\,$\downarrow$} & \textbf{Frag\,$\downarrow$} & \textbf{MTBF\,$\uparrow$} \\
\hline
\name (SAM3) & \textbf{95.8\%} & \textbf{95.8\%} & \textbf{0.0} & \textbf{6.5} & \textbf{266.3} \\
MLP-only + all losses & 78.4\% & 64.2\% & 2.0 & 16.3 & 172.9 \\
SAM3 + Hungarian & 83.5\% & 71.9\% & 7.5 & 12.8 & 183.8 \\
\hline
\end{tabular}
\vspace{-2mm}
\end{table}

Removing global self-attention reduces MOTA by 17.4 percentage points, lowers mostly-tracked trajectories from 95.8\% to 64.2\%, and increases fragmentation from 6.5 to 16.3. The MLP-only model also falls below SAM3 + Hungarian in MOTA, indicating that independent token processing cannot substitute for scene-level context in dense scenes.
\section{\textbf{Conclusion}}

We presented \name, a self-supervised framework for dense multi-camera 3D tracking that learns cross-view association and temporal identity linking from calibrated geometry alone, without identity labels, appearance-based re-identification, or ground-truth 3D trajectories. 
\name achieves 96.6\% MOTA on held-out SynFish scenes and 81.1\% on real 3D-ZeF zebrafish, outperforming strong baselines while substantially improving identity stability. Ablations show that contrastive embedding learning is essential for separating co-visible individuals, and the recovered trajectories are accurate enough to support downstream abnormal-behaviour detection (Appendix~\ref{app:anomaly}). 
The same formulation also transfers to real-world pigeons. Additional appendix experiments evaluate 30-fish scaling (Appendix~\ref{app:scalability}), audit pseudo-label noise (Appendix~\ref{app:pseudolabel_audit}), and test training-time calibration perturbations (Appendix~\ref{app:training_calibration}). The method nevertheless depends on calibration, multi-view visibility, and reliable detections. Future work could improve robustness to calibration errors and missed detections, and collect dense real-world multi-camera animal datasets with ground-truth 3D trajectories.

\section*{Acknowledgments and Disclosure of Funding}

This work was partially funded by the Research Grants Council of
Hong Kong (Ref: 17216826), the Innovation and Technology
Commission of the HKSAR Government under the ITSP-Platform
grant (Ref: ITS/335/23FP,  ITS/469/24FP), 
and a contract research project with SoftBank Corp.

\bibliographystyle{plainnat}
\bibliography{components-NIPS-revision/references}

@inproceedings{bewley2016simple,
  title={Simple online and realtime tracking},
  author={Bewley, Alex and Ge, Zongyuan and Ott, Lionel and Ramos, Fabio and Upcroft, Ben},
  booktitle={2016 IEEE international conference on image processing (ICIP)},
  pages={3464--3468},
  year={2016},
  organization={Ieee}
}

@inproceedings{naik2023pop,
  title={3d-pop-an automated annotation approach to facilitate markerless 2d-3d tracking of freely moving birds with marker-based motion capture},
  author={Naik, Hemal and Chan, Alex Hoi Hang and Yang, Junran and Delacoux, Mathilde and Couzin, Iain D and Kano, Fumihiro and Nagy, M{\'a}t{\'e}},
  booktitle={2023 IEEE/CVF Conference on Computer Vision and Pattern Recognition (CVPR)},
  pages={21274--21284},
  year={2023},
  organization={IEEE}
}

@article{wu2024cbil,
  title={Cbil: collective behavior imitation learning for fish from real videos},
  author={Wu, Yifan and Dou, Zhiyang and Ishiwaka, Yuko and Ogawa, Shun and Lou, Yuke and Wang, Wenping and Liu, Lingjie and Komura, Taku},
  journal={ACM Transactions on Graphics (TOG)},
  volume={43},
  number={6},
  pages={1--17},
  year={2024},
  publisher={ACM New York, NY, USA}
}

@article{vicsek2012collective,
  title={Collective motion},
  author={Vicsek, Tam{\'a}s and Zafeiris, Anna},
  journal={Physics reports},
  volume={517},
  number={3-4},
  pages={71--140},
  year={2012},
  publisher={Elsevier}
}

@article{cavagna2018physics,
  title={The physics of flocking: Correlation as a compass from experiments to theory},
  author={Cavagna, Andrea and Giardina, Irene and Grigera, Tom{\'a}s S},
  journal={Physics Reports},
  volume={728},
  pages={1--62},
  year={2018},
  publisher={Elsevier}
}

@article{cavagna2010scalefree,
  title={Scale-free correlations in starling flocks},
  author={Cavagna, Andrea and Cimarelli, Alessio and Giardina, Irene and Parisi, Giorgio and Santagati, Raffaele and Stefanini, Fabio and Viale, Massimiliano},
  journal={Proceedings of the National Academy of Sciences},
  volume={107},
  number={26},
  pages={11865--11870},
  year={2010},
  publisher={National Academy of Sciences}
}

@inproceedings{reynolds1987flocks,
  title={Flocks, herds and schools: A distributed behavioral model},
  author={Reynolds, Craig W},
  booktitle={Proceedings of the 14th annual conference on Computer graphics and interactive techniques},
  pages={25--34},
  year={1987}
}

@article{ballerini2008interaction,
  title={Interaction ruling animal collective behavior depends on topological rather than metric distance: Evidence from a field study},
  author={Ballerini, Michele and Cabibbo, Nicola and Candelier, Raphael and Cavagna, Andrea and Cisbani, Evaristo and Giardina, Irene and Lecomte, Vivien and Orlandi, Alberto and Parisi, Giorgio and Procaccini, Andrea and others},
  journal={Proceedings of the national academy of sciences},
  volume={105},
  number={4},
  pages={1232--1237},
  year={2008},
  publisher={National Academy of Sciences}
}

@article{couzin2002collective,
  title={Collective memory and spatial sorting in animal groups},
  author={Couzin, Iain D and Krause, Jens and James, Richard and Ruxton, Graeme D and Franks, Nigel R},
  journal={Journal of theoretical biology},
  volume={218},
  number={1},
  pages={1--11},
  year={2002},
  publisher={Elsevier}
}

@article{couzin2005effective,
  title={Effective leadership and decision-making in animal groups on the move},
  author={Couzin, Iain D and Krause, Jens and Franks, Nigel R and Levin, Simon A},
  journal={Nature},
  volume={433},
  number={7025},
  pages={513--516},
  year={2005},
  publisher={Nature Publishing Group UK London}
}

@article{heins2024collective,
  title={Collective behavior from surprise minimization},
  author={Heins, Conor and Millidge, Beren and Da Costa, Lancelot and Mann, Richard P and Friston, Karl J and Couzin, Iain D},
  journal={Proceedings of the National Academy of Sciences},
  volume={121},
  number={17},
  pages={e2320239121},
  year={2024},
  publisher={National Academy of Sciences}
}

@article{dell2014automated,
  title={Automated image-based tracking and its application in ecology},
  author={Dell, Anthony I and Bender, John A and Branson, Kristin and Couzin, Iain D and de Polavieja, Gonzalo G and Noldus, Lucas PJJ and P{\'e}rez-Escudero, Alfonso and Perona, Pietro and Straw, Andrew D and Wikelski, Martin and others},
  journal={Trends in ecology \& evolution},
  volume={29},
  number={7},
  pages={417--428},
  year={2014},
  publisher={Elsevier}
}

@article{hofmann2014evolutionary,
  title={An evolutionary framework for studying mechanisms of social behavior},
  author={Hofmann, Hans A and Beery, Annaliese K and Blumstein, Daniel T and Couzin, Iain D and Earley, Ryan L and Hayes, Loren D and Hurd, Peter L and Lacey, Eileen A and Phelps, Steven M and Solomon, Nancy G and others},
  journal={Trends in ecology \& evolution},
  volume={29},
  number={10},
  pages={581--589},
  year={2014},
  publisher={Elsevier}
}

@article{perezescudero2014idtracker,
author = {Pérez-Escudero, Alfonso and Vicente-Page, Julián and Hinz, Robert and Arganda, Sara and Polavieja, Gonzalo},
year = {2014},
month = {06},
pages = {},
title = {IdTracker: Tracking individuals in a group by automatic identification of unmarked animals},
volume = {11},
journal = {Nature methods},
doi = {10.1038/nmeth.2994}
}

@article{romeroferrero2019idtrackerai,
author = {Romero-Ferrero, Francisco and Bergomi, Mattia and Hinz, Robert and Heras, Francisco and Polavieja, Gonzalo},
year = {2019},
month = {02},
pages = {},
title = {idtracker.ai: Tracking all individuals in large collectives of unmarked animals},
volume = {16},
journal = {Nature Methods},
doi = {10.1038/s41592-018-0295-5}
}

@article{perezescudero2013collective,
    doi = {10.1371/journal.pcbi.1002282},
    author = {Pérez-Escudero, Alfonso AND de Polavieja, Gonzalo G.},
    journal = {PLOS Computational Biology},
    publisher = {Public Library of Science},
    title = {Collective Animal Behavior from Bayesian Estimation and Probability Matching},
    year = {2011},
    month = {11},
    volume = {7},
    url = {https://doi.org/10.1371/journal.pcbi.1002282},
    pages = {1-14},
    number = {11},

}

@article{herbertread2011inferring,
  title={Inferring the rules of interaction of shoaling fish},
  author={Herbert-Read, James E and Perna, Andrea and Mann, Richard P and Schaerf, Timothy M and Sumpter, David JT and Ward, Ashley JW},
  journal={Proceedings of the National Academy of Sciences},
  volume={108},
  number={46},
  pages={18726--18731},
  year={2011},
  publisher={National Academy of Sciences}
}

@article{calovi2015collective,
  title={Collective response to perturbations in a data-driven fish school model},
  author={Calovi, Daniel S and Lopez, Ugo and Schuhmacher, Paul and Chat{\'e}, Hugues and Sire, Cl{\'e}ment and Theraulaz, Guy},
  journal={Journal of The Royal Society Interface},
  volume={12},
  number={104},
  pages={20141362},
  year={2015}
}

@article{newbolt2019flow,
  title={Flow interactions between uncoordinated flapping swimmers give rise to group cohesion},
  author={Newbolt, Joel W and Zhang, Jun and Ristroph, Leif},
  journal={Proceedings of the National Academy of Sciences},
  volume={116},
  number={7},
  pages={2419--2424},
  year={2019},
  publisher={National Academy of Sciences}
}

@article{aoki1982simulation,
author = {Ichiro AOKI. },
title = {A simulation study on the schooling mechanism in fish.},
journal = {NIPPON SUISAN GAKKAISHI},
year = 1982,
volume = {48},
number = {8},
pages = {1081-1088},
doi = {10.2331/suisan.48.1081}
}

@article{filella2018model,
  title={Model of collective fish behavior with hydrodynamic interactions},
  author={Filella, Audrey and Nadal, Fran{\c{c}}ois and Sire, Cl{\'e}ment and Kanso, Eva and Eloy, Christophe},
  journal={Physical review letters},
  volume={120},
  number={19},
  pages={198101},
  year={2018},
  publisher={APS}
}

@article{niwa1996newtonian,
  title={Newtonian dynamical approach to fish schooling},
  author={Niwa, Hiro-Sato},
  journal={Journal of Theoretical Biology},
  volume={181},
  number={1},
  pages={47--63},
  year={1996},
  publisher={Elsevier}
}

@article{ispolatov2016collective,
  title={Computing in fish schools},
  author={Ispolatov, Yaroslav},
  journal={ELife},
  volume={5},
  pages={e12852},
  year={2016},
  publisher={eLife Sciences Publications, Ltd}
}

@article{jiang2023collective,
  title={Collective motions of fish originate from balanced local perceptual interactions and individual stochastics},
  author={Jiang, Mingjie and Zhou, Anyu and Chen, Runping and Yang, Yuqin and Dong, Hao and Wang, Wei},
  journal={Physical Review E},
  volume={107},
  number={2},
  pages={024411},
  year={2023},
  publisher={APS}
}

@article{chung2018survey,
  title={A survey on aerial swarm robotics},
  author={Chung, Soon-Jo and Paranjape, Aditya Avinash and Dames, Philip and Shen, Shaojie and Kumar, Vijay},
  journal={IEEE Transactions on robotics},
  volume={34},
  number={4},
  pages={837--855},
  year={2018},
  publisher={IEEE}
}

@article{zhou2022swarm,
  title={Swarm of micro flying robots in the wild},
  author={Zhou, Xin and Wen, Xiangyong and Wang, Zhepei and Gao, Yuman and Li, Haojia and Wang, Qianhao and Yang, Tiankai and Lu, Haojian and Cao, Yanjun and Xu, Chao and others},
  journal={Science robotics},
  volume={7},
  number={66},
  pages={eabm5954},
  year={2022},
  publisher={American Association for the Advancement of Science}
}

@inproceedings{samba2024,
  title={Samba: Synchronized set-of-sequences modeling for multiple object tracking},
  author={Segu, Mattia and Piccinelli, Luigi and Li, Siyuan and Yang, Yung-Hsu and Van Gool, Luc and Schiele, Bernt},
  booktitle={International Conference on Learning Representations},
  volume={2025},
  pages={30057--30070},
  year={2025}
}

@inproceedings{motip2024,
  title={Multiple object tracking as id prediction},
  author={Gao, Ruopeng and Qi, Ji and Wang, Limin},
  booktitle={2025 IEEE/CVF Conference on Computer Vision and Pattern Recognition (CVPR)},
  pages={27883--27893},
  year={2025},
  organization={IEEE}
}

@inproceedings{trackformer2022,
  title={Trackformer: Multi-object tracking with transformers},
  author={Meinhardt, Tim and Kirillov, Alexander and Leal-Taixe, Laura and Feichtenhofer, Christoph},
  booktitle={2022 IEEE/CVF conference on computer vision and pattern recognition (CVPR)},
  pages={8834--8844},
  year={2022},
  organization={IEEE}
}

@inproceedings{memot2022,
  title={Memot: Multi-object tracking with memory},
  author={Cai, Jiarui and Xu, Mingze and Li, Wei and Xiong, Yuanjun and Xia, Wei and Tu, Zhuowen and Soatto, Stefano},
  booktitle={2022 IEEE/CVF Conference on Computer Vision and Pattern Recognition (CVPR)},
  pages={8080--8090},
  year={2022},
  organization={IEEE}
}

@inproceedings{cotracker2024,
  title={Cotracker: It is better to track together},
  author={Karaev, Nikita and Rocco, Ignacio and Graham, Benjamin and Neverova, Natalia and Vedaldi, Andrea and Rupprecht, Christian},
  booktitle={European conference on computer vision},
  pages={18--35},
  year={2024},
  organization={Springer}
}

@inproceedings{rest2023,
  title={Rest: A reconfigurable spatial-temporal graph model for multi-camera multi-object tracking},
  author={Cheng, Cheng-Che and Qiu, Min-Xuan and Chiang, Chen-Kuo and Lai, Shang-Hong},
  booktitle={2023 IEEE/CVF International Conference on Computer Vision (ICCV)},
  pages={10017--10026},
  year={2023},
  organization={IEEE}
}

@inproceedings{mctr2024,
  title={Mctr: Multi camera tracking transformer},
  author={Niculescu-Mizil, Alexandru and Patel, Deep and Melvin, Iain},
  booktitle={2025 IEEE/CVF Winter Conference on Applications of Computer Vision Workshops (WACVW)},
  pages={816--826},
  year={2025},
  organization={IEEE}
}

@article {adatrack2024,
	Title = {ADA-Track++: End-to-End Multi-Camera 3D Multi-Object Tracking With Alternating Detection and Association},
	Author = {Ding, Shuxiao and Schneider, Lukas and Cordts, Marius and Gall, Juergen},
	DOI = {10.1109/tpami.2025.3613269},
	Number = {1},
	Volume = {48},
	Month = {January},
	Year = {2026},
	Journal = {IEEE transactions on pattern analysis and machine intelligence},
	ISSN = {0162-8828},
	Pages = {482—499},
	URL = {https://doi.org/10.1109/TPAMI.2025.3613269},
}

@inproceedings{lmgp2022,
  title={Lmgp: Lifted multicut meets geometry projections for multi-camera multi-object tracking},
  author={Nguyen, Duy MH and Henschel, Roberto and Rosenhahn, Bodo and Sonntag, Daniel and Swoboda, Paul},
  booktitle={2022 IEEE/CVF Conference on Computer Vision and Pattern Recognition (CVPR)},
  pages={8856--8865},
  year={2022},
  organization={IEEE}
}

@inproceedings{dyglip2021,
  title={Dyglip: A dynamic graph model with link prediction for accurate multi-camera multiple object tracking},
  author={Quach, Kha Gia and Nguyen, Pha and Le, Huu and Truong, Thanh-Dat and Duong, Chi Nhan and Tran, Minh-Triet and Luu, Khoa},
  booktitle={2021 IEEE/CVF Conference on Computer Vision and Pattern Recognition (CVPR)},
  pages={13779--13788},
  year={2021},
  organization={IEEE}
}

@ARTICLE{tracta2020,
  author={He, Yuhang and Wei, Xing and Hong, Xiaopeng and Shi, Weiwei and Gong, Yihong},
  journal={IEEE Transactions on Image Processing}, 
  title={Multi-Target Multi-Camera Tracking by Tracklet-to-Target Assignment}, 
  year={2020},
  volume={29},
  number={},
  pages={5191-5205},
  doi={10.1109/TIP.2020.2980070}}

@inproceedings{messytable2020,
  title={Messytable: Instance association in multiple camera views},
  author={Cai, Zhongang and Zhang, Junzhe and Ren, Daxuan and Yu, Cunjun and Zhao, Haiyu and Yi, Shuai and Yeo, Chai Kiat and Change Loy, Chen},
  booktitle={European Conference on Computer Vision},
  pages={1--16},
  year={2020},
  organization={Springer}
}

@inproceedings{vitp3de2023,
  title={{ViT-P3DE}*: Vision Transformer Based Multi-Camera Instance Association with Pseudo {3D} Position Embedding},
  author={Seo, Minseok and Lee, Hyuk-Jae and Nguyen, Xuan Truong},
  booktitle={IJCAI},
  pages={1340--1350},
  year={2023}
}

@inproceedings{selfmva2025,
  title={Learning from synchronization: self-supervised uncalibrated multi-view person association in challenging scenes},
  author={Chen, Keqi and Srivastav, Vinkle and Mutter, Didier and Padoy, Nicolas},
  booktitle={2025 IEEE/CVF Conference on Computer Vision and Pattern Recognition (CVPR)},
  pages={24419--24428},
  year={2025},
  organization={IEEE}
}

@inproceedings{mvmhat2021,
  title={Self-supervised multi-view multi-human association and tracking},
  author={Gan, Yiyang and Han, Ruize and Yin, Liqiang and Feng, Wei and Wang, Song},
  booktitle={Proceedings of the 29th ACM international conference on multimedia},
  pages={282--290},
  year={2021}
}

@article{vo2020selfsupervised,
  title={Self-supervised multi-view person association and its applications},
  author={Vo, Minh and Yumer, Ersin and Sunkavalli, Kalyan and Hadap, Sunil and Sheikh, Yaser and Narasimhan, Srinivasa G},
  journal={IEEE transactions on pattern analysis and machine intelligence},
  volume={43},
  number={8},
  pages={2794--2808},
  year={2020},
  publisher={IEEE}
}

@article{yolo26,
  title={Ultralytics YOLO26: unified real-time end-to-end vision models},
  author={Jocher, Glenn and Qiu, Jing and Liu, Mengyu and Lyu, Shuai and Akyon, Fatih Cagatay and Kalfaoglu, Muhammet Esat},
  journal={arXiv preprint arXiv:2606.03748},
  year={2026}
}

@inproceedings{sam2,
  title={Sam 2: Segment anything in images and videos},
  author={Ravi, Nikhila and Gabeur, Valentin and Hu, Yuan-Ting and Hu, Ronghang and Ryali, Chaitanya and Ma, Tengyu and Khedr, Haitham and R{\"a}dle, Roman and Rolland, Chloe and Gustafson, Laura and others},
  booktitle={International Conference on Learning Representations},
  volume={2025},
  pages={28085--28128},
  year={2025}
}

@inproceedings{sam3,
  title={Sam 3: Segment anything with concepts},
  author={Carion, Nicolas and Gustafson, Laura and Hu, Yuan-Ting and Debnath, Shoubhik and Hu, Ronghang and Suris Coll-Vinent, Didac and Ryali, Chaitanya and Alwala, Kalyan Vasudev and Khedr, Haitham and Huang, Andrew and others},
  booktitle={International conference on learning representations},
  volume={2026},
  pages={138846--138923},
  year={2026}
}

@article{kuhn1955hungarian,
  title={The Hungarian method for the assignment problem},
  author={Kuhn, Harold W},
  journal={Naval research logistics quarterly},
  volume={2},
  number={1-2},
  pages={83--97},
  year={1955},
  publisher={Wiley Online Library}
}

@article{hartley1997triangulation,
  title={Triangulation},
  author={Hartley, Richard I and Sturm, Peter},
  journal={Computer vision and image understanding},
  volume={68},
  number={2},
  pages={146--157},
  year={1997},
  publisher={Elsevier}
}

@book{hartley2003multiview,
  title={Multiple view geometry in computer vision},
  author={Hartley, Richard and Zisserman, Andrew and others},
  volume={2},
  number={3},
  year={2003},
  publisher={Cambridge university press Cambridge}
}

@inproceedings{nair2010relu,
  title={Rectified linear units improve restricted boltzmann machines},
  author={Nair, Vinod and Hinton, Geoffrey E},
  booktitle={Proceedings of the 27th international conference on machine learning (ICML-10)},
  pages={807--814},
  year={2010}
}

@article{ba2016layernorm,
  title={Layer normalization},
  author={Ba, Jimmy Lei and Kiros, Jamie Ryan and Hinton, Geoffrey E},
  journal={arXiv preprint arXiv:1607.06450},
  year={2016}
}

@inproceedings{rope2024,
  title={Rotary position embedding for vision transformer},
  author={Heo, Byeongho and Park, Song and Han, Dongyoon and Yun, Sangdoo},
  booktitle={European Conference on Computer Vision},
  pages={289--305},
  year={2024},
  organization={Springer}
}

@inproceedings{vggt2025,
  title={Vggt: Visual geometry grounded transformer},
  author={Wang, Jianyuan and Chen, Minghao and Karaev, Nikita and Vedaldi, Andrea and Rupprecht, Christian and Novotny, David},
  booktitle={2025 IEEE/CVF Conference on Computer Vision and Pattern Recognition (CVPR)},
  pages={5294--5306},
  year={2025},
  organization={IEEE}
}

@inproceedings{pedersen2020zef,
  title={3d-zef: A 3d zebrafish tracking benchmark dataset},
  author={Pedersen, Malte and Haurum, Joakim Bruslund and Bengtson, Stefan Hein and Moeslund, Thomas B},
  booktitle={2020 IEEE/CVF Conference on Computer Vision and Pattern Recognition (CVPR)},
  pages={2423--2433},
  year={2020},
  organization={IEEE}
}

@article{oord2018infonce,
  title={Representation learning with contrastive predictive coding},
  author={Oord, Aaron van den and Li, Yazhe and Vinyals, Oriol},
  journal={arXiv preprint arXiv:1807.03748},
  year={2018}
}

@article{dinov3,
  title={Dinov3},
  author={Sim{\'e}oni, Oriane and Vo, Huy V and Seitzer, Maximilian and Baldassarre, Federico and Oquab, Maxime and Jose, Cijo and Khalidov, Vasil and Szafraniec, Marc and Yi, Seungeun and Ramamonjisoa, Micha{\"e}l and others},
  journal={arXiv preprint arXiv:2508.10104},
  year={2025}
}

@inproceedings{chong2017abnormal,
  title={Abnormal event detection in videos using spatiotemporal autoencoder},
  author={Chong, Yong Shean and Tay, Yong Haur},
  booktitle={International symposium on neural networks},
  pages={189--196},
  year={2017},
  organization={Springer}
}

@article{martins2012behavioural,
  title={Behavioural indicators of welfare in farmed fish},
  author={Martins, Catarina IM and Galhardo, Leonor and Noble, Chris and Damsg{\aa}rd, B{\o}rge and Spedicato, Maria T and Zupa, Walter and Beauchaud, Marilyn and Kulczykowska, Ewa and Massabuau, Jean-Charles and Carter, Toby and others},
  journal={Fish Physiology and Biochemistry},
  volume={38},
  number={1},
  pages={17--41},
  year={2012},
  publisher={Springer}
}

@article{papadakis2012anomaly,
  title={A computer-vision system and methodology for the analysis of fish behavior},
  author={Papadakis, Vassilis M and Papadakis, Ioannis E and Lamprianidou, Fani and Glaropoulos, Alexios and Kentouri, Maroudio},
  journal={Aquacultural engineering},
  volume={46},
  pages={53--59},
  year={2012},
  publisher={Elsevier}
}

@article{li2019gan_anomaly,
  title={Anomaly detection with generative adversarial networks for multivariate time series},
  author={Li, Dan and Chen, Dacheng and Goh, Jonathan and Ng, See-kiong},
  journal={arXiv preprint arXiv:1809.04758},
  year={2018}
}

@inproceedings{liu2008isolation,
  title={Isolation forest},
  author={Liu, Fei Tony and Ting, Kai Ming and Zhou, Zhi-Hua},
  booktitle={2008 eighth ieee international conference on data mining},
  pages={413--422},
  year={2008},
  organization={Ieee}
}

@inproceedings{breunig2000lof,
  title={LOF: identifying density-based local outliers},
  author={Breunig, Markus M and Kriegel, Hans-Peter and Ng, Raymond T and Sander, J{\"o}rg},
  booktitle={Proceedings of the 2000 ACM SIGMOD international conference on Management of data},
  pages={93--104},
  year={2000}
}

\newpage
\appendices

\section*{Appendix Contents}

\begin{tabular}{ll}
\textbf{Appendix} & \textbf{Page} \\
\hline
A. \name Training Algorithm & \pageref{app:algorithm} \\
B. Network Architecture & \pageref{app:architecture} \\
C. Training Details & \pageref{app:training} \\
D. Inference Implementation & \pageref{app:inference} \\
E. Dataset Statistics & \pageref{app:dataset_stats} \\
F. Camera Calibration Parameters & \pageref{app:camera_params} \\
G. Per-Scene Results: SynFish Test Set & \pageref{app:ue5_perscene} \\
H. Per-Scene Results: 3D-ZeF Test Set & \pageref{app:zebra_perscene} \\
I. Temporal Linking Weight Ablation & \pageref{app:temporal_weight_ablation} \\
J. Robustness to Camera Calibration Noise & \pageref{app:robustness} \\
K. Robustness to Detection Noise & \pageref{app:det_robustness} \\
L. Runtime Performance & \pageref{app:runtime} \\
M. Extended Method & \pageref{app:extended_method} \\
N. Extended Experiment & \pageref{app:extended_experiment} \\
O. Proof-of-principle downstream analysis on Abnormal Behaviour Detection & \pageref{app:anomaly} \\
P. Cross-species generalisation to real-world pigeons & \pageref{app:pigeon} \\
Q. 3D-SORT Kinematic Baseline & \pageref{app:sort_baseline} \\
R. Density and Computational Scaling & \pageref{app:scalability} \\
S. Pseudo-Label Noise Audit & \pageref{app:pseudolabel_audit} \\
T. Training-Time Calibration Noise & \pageref{app:training_calibration} \\
\end{tabular}

\clearpage

\section{\name Training Algorithm}\label{app:algorithm}

\begin{algorithm}[H]
\small
\DontPrintSemicolon
\SetAlgoLined
\SetAlgoNlRelativeSize{-1}
\caption{\name Training}
\label{alg:densefish}
\KwIn{Scenes $\mathcal{S}$ with calibrated cameras $\{K_c, R_c, t_c\}_{c=1}^{C}$ and 2D detections}
\KwOut{Encoder $\phi$, association transformer $\psi$, temporal predictor $\rho$}
\For{\textnormal{epoch} $= 1, \ldots, E$}{
  $\texttt{full} \leftarrow (\text{epoch} > E_{\mathrm{warmup}})$ \;
  \For{\textnormal{each scene} $\mathcal{S}_k$}{
    $\mathcal{M} \leftarrow \emptyset$ \tcp*{track memory (sliding window)}
    \For{\textnormal{each consecutive pair} $(t,\, t{+}1)$}{
      \tcp{Stage 1: Encode \& Associate (Sec.~3.3--3.4)}
    $F_\bullet \leftarrow \phi(\text{dets}_\bullet, \text{cams})$;\quad $H_\bullet \leftarrow \psi_{\mathrm{tf}}(\textsc{RoPE}(\psi_{\mathrm{proj}}(F_\bullet)))$ \quad for $\bullet \in \{t, t{+}1\}$ \;
      $P_\bullet \leftarrow \sigma\bigl(\psi_{\mathrm{pw}}([H_\bullet^i \| H_\bullet^j])\bigr)\;\forall\; c_i \neq c_j$ \;
      \tcp{Stage 2: Pseudo-Supervision \& Association Loss (Sec.~3.5--3.6)}
      $(G_\bullet, C_\bullet) \leftarrow \textsc{PseudoGT}(\text{dets}_\bullet, \text{cams})$ \;
      $\mathcal{L}_{\mathrm{asso}} \leftarrow \sum_\bullet \textsc{WeightedBCE}(P_\bullet, G_\bullet, C_\bullet)$ \;
      \eIf{\textnormal{\texttt{full}}}{
        \tcp{Stage 3: Grouping \& Triangulation (Sec.~3.7)}
        $A_\bullet \leftarrow 0.6 P_\bullet + 0.4\,\frac{\cos(H_\bullet, H_\bullet)+1}{2}$ \;
        $\mathcal{G}_\bullet \leftarrow \textsc{GroupTriangulate}(A_\bullet, \text{dets}_\bullet, \text{cams})$;\; prune by confidence \;
        $\bar{\boldsymbol{h}}_g \leftarrow \ell_2\text{-norm}(\tfrac{1}{|g|}\sum_{i\in g}\boldsymbol{h}_i)$ for each group $g$ \;
        \tcp{Stage 4: Contrastive Learning (Sec.~3.8)}
        $\mathcal{L}_{\mathrm{ctr}} \leftarrow \sum_\bullet \textsc{InfoNCE}(H_\bullet, \{(i,j) \mid i,j \in g\}, \tau)$ \;
        \tcp{Stage 5: Temporal Linking \& Prediction (Sec.~3.9)}
        $\mathcal{T} \leftarrow \textsc{Hungarian}\bigl(\alpha\, d_{\mathrm{3D}} + \beta\,(1{-}\cos)\bigr)$ gated by thresholds \;
        $\mathcal{T}_{\mathrm{fb}} \leftarrow \textsc{FallbackMatch}(\mathcal{G}_{t+1}, \mathcal{M})$ \;
        $\mathcal{L}_{\mathrm{temp}} \leftarrow \textsc{WeightedMSE}\bigl(\rho(\bar{\boldsymbol{h}}_{g_t}),\; \bar{\boldsymbol{h}}_{g_{t+1}},\; \mathcal{T} \cup \mathcal{T}_{\mathrm{fb}}\bigr)$ \;
      }{
        $\mathcal{L}_{\mathrm{ctr}}, \mathcal{L}_{\mathrm{temp}} \leftarrow 0$ \tcp*{warmup: $\mathcal{L}_{\mathrm{asso}}$ only}
      }
      $\mathcal{L} \leftarrow \lambda_{\mathrm{asso}} \mathcal{L}_{\mathrm{asso}} + \lambda_{\mathrm{ctr}} \mathcal{L}_{\mathrm{ctr}} + \lambda_{\mathrm{temp}} \mathcal{L}_{\mathrm{temp}}$ \;
      Update $\phi, \psi, \rho$ via AdamW (grad clip); append $\mathcal{G}_t$ to $\mathcal{M}$ \;
    }
  }
}
\end{algorithm}

\section{Network Architecture}\label{app:architecture}

\name consists of three learned modules: a \emph{Geometric Feature Encoder}~$\phi$, an \emph{Association Transformer}~$\psi$, and a \emph{Temporal Predictor}~$\rho$.
All modules are trained jointly end-to-end.
This section provides the full architectural specification of each component, including layer dimensions and parameter counts.

\paragraph{Geometric Feature Encoder $\phi$.}\label{app:arch_encoder}

The encoder maps each 2D detection into a geometric embedding that captures its spatial location, scale, camera assignment, detection confidence, and world-space ray direction.
For each detection with bounding box $(x_1,y_1,x_2,y_2)$, segmentation centroid $(c_x,c_y)$, camera ID $c$, and detection score $s$, we construct a 9-dimensional raw feature vector $f_i^{\mathrm{raw}}$:
\begin{equation}
    f_i^{\mathrm{raw}} = \bigl[\,\tfrac{c_x}{W},\;\tfrac{c_y}{H},\;\tfrac{x_2{-}x_1}{W},\;\tfrac{y_2{-}y_1}{H},\;\tfrac{c}{C{-}1},\;s,\;r_x,\;r_y,\;r_z\,\bigr]
\end{equation}
where $(W,H)$ is the image resolution, $C$ is the number of cameras, and $(r_x,r_y,r_z)$ is the unit-length world-space ray obtained by unprojecting the centroid through the camera intrinsic inverse $K^{-1}$ and rotating by $R^\top$.

This 9-D vector is processed by a 2-layer MLP with a hidden dimension of $64$, followed by LayerNorm, producing a $d{=}128$ dimensional geometric embedding $\boldsymbol{f}_i = \phi(f_i^{\mathrm{raw}})$:
\begin{equation}
    \boldsymbol{f}_i = \text{LayerNorm}\bigl(\text{Linear}_{64 \to 128}\bigl(\text{ReLU}\bigl(\text{Linear}_{9 \to 64}(f_i^{\mathrm{raw}})\bigr)\bigr)\bigr) \in \mathbb{R}^{128}
\end{equation}

\noindent\textbf{Parameter count:}
$\text{Linear}_{9 \to 64}$: $9 \times 64 + 64 = 640$;
$\text{Linear}_{64 \to 128}$: $64 \times 128 + 128 = 8{,}320$;
$\text{LayerNorm}_{128}$: $2 \times 128 = 256$.
\textbf{Total encoder MLP:} $9{,}216$ parameters.

\paragraph{Association Transformer $\psi$.}\label{app:arch_assoc}

The association transformer takes the set of per-detection embeddings from $\phi$ and produces both contextualized embeddings $\mathbf{h}_i$ and pairwise cross-camera association probabilities~$P_{ij}$.
It consists of three sub-modules:

\paragraph{Input projection.}
A single linear layer projects the encoder output to the transformer dimension:
$\psi_{\mathrm{proj}}: \mathbb{R}^{d_{\mathrm{in}}} \to \mathbb{R}^{128}$ where $d_{\mathrm{in}} = 128$ (geometry-only mode).

\paragraph{2D Rotary Position Embedding (RoPE).}
After input projection and before entering the transformer layers, projected detection tokens receive 2D RoPE~\cite{rope2024} based on their normalised bounding-box centres $(c_y, c_x) \in [0,1]^2$.
The 128-dimensional embedding is split into two 64-dimensional halves: the first half encodes the vertical position $c_y$ and the second half encodes the horizontal position $c_x$.
Within each half, we form 32 pairs and apply the standard rotary transform:
\begin{equation}
    \theta_k = \frac{c_\text{axis}}{10000^{2k/64}}\cdot\pi, \quad k = 0, \ldots, 31
\end{equation}
\begin{equation}
    (x_1', x_2') = (x_1 \cos\theta_k - x_2 \sin\theta_k,\; x_1 \sin\theta_k + x_2 \cos\theta_k)
\end{equation}
RoPE has no learnable parameters; it enables the transformer to reason about detection spatial relationships in a translation-equivariant manner.

\paragraph{Transformer encoder.}
A standard 4-layer Transformer encoder with pre-norm architecture:
\begin{itemize}[nosep]
    \item Layers: 4, attention heads: 4, head dimension: 32
    \item Model dimension $d = 128$, feed-forward dimension $d_{\mathrm{ff}} = 256$
    \item Dropout: 0.1, activation: ReLU
    \item Pre-LayerNorm (norm-first) variant
\end{itemize}
Each layer applies multi-head self-attention followed by a position-wise feed-forward network.
The transformer operates on the full set of $N$ detection tokens from all cameras simultaneously, enabling cross-camera reasoning.

\textbf{Per-layer parameter count:}
Self-attention QKV: $3 \times (128 \times 128 + 128) = 49{,}536$;
Output projection: $128 \times 128 + 128 = 16{,}512$;
FFN: $(128 \times 256 + 256) + (256 \times 128 + 128) = 65{,}920$;
LayerNorm $\times 2$: $2 \times 256 = 512$.
\textbf{Per layer: $132{,}480$.  Four layers: $529{,}920$.}

\paragraph{Pairwise MLP.}
For every pair of detections $(i, j)$ from \emph{different} cameras ($c_i \neq c_j$), we concatenate their transformer outputs and predict an association probability:
\begin{equation}
    P_{ij} = \sigma\!\bigl(\text{Linear}_{128 \to 1}\bigl(\text{ReLU}\bigl(\text{Linear}_{256 \to 128}([\mathbf{h}_i \| \mathbf{h}_j])\bigr)\bigr)\bigr)
\end{equation}
\textbf{Pairwise MLP parameters:} $(256 \times 128 + 128) + (128 \times 1 + 1) = 33{,}025$.

\textbf{Total association transformer:}
Input projection ($16{,}512$) + transformer layers ($529{,}920$) + pairwise MLP ($33{,}025$) = \textbf{$\approx$579K} parameters.
Note that 2D RoPE contributes \textbf{zero learnable parameters}---it stores only a fixed frequency buffer computed from deterministic sinusoidal formulas and applies parameter-free rotations to the embeddings.

\paragraph{Temporal Predictor $\rho$.}\label{app:arch_predictor}

The temporal predictor is a small MLP that maps a group embedding $\bar{\boldsymbol{h}}_g$ at frame $t$ to its predicted embedding at frame $t{+}1$:
\begin{equation}
    \rho(\bar{\boldsymbol{h}}_g) = \text{Linear}_{256 \to 128}\bigl(\text{LayerNorm}_{256}\bigl(\text{ReLU}\bigl(\text{Linear}_{128 \to 256}(\bar{\boldsymbol{h}}_g)\bigr)\bigr)\bigr) \in \mathbb{R}^{128}
\end{equation}

\textbf{Parameter count:}
$\text{Linear}_{128 \to 256}$: $128 \times 256 + 256 = 33{,}024$;
$\text{LayerNorm}_{256}$: $2 \times 256 = 512$;
$\text{Linear}_{256 \to 128}$: $256 \times 128 + 128 = 32{,}896$.
\textbf{Total predictor: $66{,}432$} parameters.

At inference, when a track has been unobserved for $g$ frames, $\rho$ is applied iteratively (up to~5 times) to roll the embedding forward, approximating what the track's representation would look like after $g$ frames of motion:
$\hat{\bar{\boldsymbol{h}}}^{(t+g)} = \rho^{(g)}(\bar{\boldsymbol{h}}^{(t)})$.

\paragraph{Total model size.}\label{app:arch_total}

\begin{table}[H]
\centering
\caption{\name parameter budget (geometry-only mode, no visual backbone).}
\label{tab:param_budget}
\small
\begin{tabular}{lrr}
\hline
\textbf{Module} & \textbf{Parameters} & \textbf{\%} \\
\hline
Geometric Feature Encoder $\phi$ & 9{,}216 & 1.4\% \\
Association Transformer $\psi$ & 579{,}457 & 88.5\% \\
Temporal Predictor $\rho$ & 66{,}432 & 10.1\% \\
\hline
\textbf{Total (trainable)} & \textbf{$\approx$655K} & 100\% \\
\hline
\end{tabular}
\end{table}

\noindent The entire \name model has approximately \textbf{655K trainable parameters}---orders of magnitude smaller than typical vision transformer trackers ($>$50M parameters).
This compact architecture is enabled by the geometry-only design: since all fish are visually indistinguishable, a large visual backbone is unnecessary.

\section{Training Details}\label{app:training}

\paragraph{Training protocol.}\label{app:train_protocol}

\name is trained for \textbf{400 epochs} with a \textbf{100-epoch warmup} on the SynFish training set (8 scenes, 4--16 fish, $\sim$350 frames each).
During the warmup phase (epochs 1--100), only the association loss $\mathcal{L}_{\mathrm{asso}}$ is active; grouping, contrastive learning, and temporal linking are disabled.
This allows the encoder and association transformer to learn basic cross-camera correspondence before the more demanding grouping and tracking losses are introduced.
After warmup (epochs 101--400), all three losses ($\mathcal{L}_{\mathrm{asso}}$, $\mathcal{L}_{\mathrm{ctr}}$, $\mathcal{L}_{\mathrm{temp}}$) are jointly optimised.

Training iterates over scenes sequentially within each epoch; within each scene, consecutive frame pairs $(t, t{+}1)$ are processed in temporal order.
This sequential processing is essential because the temporal prediction loss requires a sliding-window track memory ($W{=}10$ past frames) that accumulates as the sequence progresses.
Each frame pair constitutes one gradient step: the total loss is computed and backpropagated, and all parameters are updated via AdamW before proceeding to the next pair.

\paragraph{Optimiser and hyperparameters.}\label{app:train_hyper}

\begin{table}[H]
\centering
\caption{Complete \name training hyperparameters.}
\label{tab:hyperparams}
\small
\begin{tabular}{llr}
\hline
\textbf{Category} & \textbf{Hyperparameter} & \textbf{Value} \\
\hline
\multirow{4}{*}{Optimiser}
& Algorithm & AdamW \\
& Learning rate & $1 \times 10^{-4}$ \\
& Weight decay & $1 \times 10^{-4}$ \\
& Gradient clipping (max norm) & 5.0 \\
\hline
\multirow{3}{*}{Loss weights}
& $\lambda_{\mathrm{asso}}$ (association) & 1.0 \\
& $\lambda_{\mathrm{ctr}}$ (contrastive) & 0.5 \\
& $\lambda_{\mathrm{temp}}$ (temporal) & 0.25 \\
\hline
\multirow{4}{*}{Grouping}
& Association threshold $\theta_{\mathrm{assoc}}$ & 0.6 \\
& Affinity weights ($w_P$, $w_F$) & (0.6, 0.4) \\
& Minimum group confidence $\theta_{\mathrm{conf}}$ & 0.3 \\
& Two-camera acceptance threshold & 0.8 \\
\hline
\multirow{4}{*}{Temporal linking}
& Distance weight $\alpha$ & 0.6 \\
& Appearance weight $\beta$ & 0.4 \\
& Distance threshold $\theta_{\mathrm{dist}}$ & 0.5 \\
& Similarity threshold $\theta_{\mathrm{sim}}$ & 0.6 \\
\hline
\multirow{2}{*}{Fallback matching}
& Gap distance scale & 0.15 \\
& Gap similarity drop & 0.04 \\
\hline
\multirow{2}{*}{Post-hoc merge}
& Merge distance threshold $\theta_{\mathrm{merge}}$ & 2.0 \\
& Merge similarity threshold & 0.45 \\
\hline
\multirow{2}{*}{Contrastive}
& Temperature $\tau$ & 0.07 \\
& InfoNCE & detection-level \\
\hline
\multirow{3}{*}{Schedule}
& Total epochs & 400 \\
& Warmup epochs $E_{\mathrm{warmup}}$ & 100 \\
& Track memory window $W$ & 10 \\
\hline
\end{tabular}
\end{table}

\paragraph{Pseudo ground-truth caching.}\label{app:train_pseudogt}

Pseudo ground-truth labels $(G, C)$ are computed by pairwise triangulation and bidirectional reprojection-containment checks (see Sec.~3.5), then cached to disk as compressed \texttt{.npz} files before the first epoch.
This one-time precomputation eliminates redundant geometric computation during training and ensures deterministic supervision across epochs.
For each frame, the cached files store the binary assignment matrix $G \in \{-1,0,1\}^{N \times N}$ and the continuous confidence matrix $C \in [0,1]^{N \times N}$.

\paragraph{Feature reuse across frame pairs.}\label{app:train_reuse}

To reduce redundant computation, when processing consecutive frame pairs $(t,t{+}1)$ and $(t{+}1,t{+}2)$, the encoder output for frame $t{+}1$ computed in the first pair is \emph{detached} from the computational graph and reused as the frame~$t$ input for the next pair.
This halves the per-scene encoding cost with no effect on gradient quality (since only the current pair's computation graph is retained for backpropagation).

\paragraph{Checkpoint selection.}\label{app:train_ckpt}

Checkpoints are saved every 100 epochs and at the final epoch.
For the reported SAM3 and YOLO26+SAM2 models, \texttt{best\_ckpt.pth} is selected by the lowest training objective; the resulting checkpoints are from epochs 396 and 397, respectively.
The held-out split is therefore not used for model selection.

\section{Inference Implementation}\label{app:inference}

At inference, \name processes a scene in three sequential passes followed by post-processing.
The inference pipeline uses the trained encoder, association transformer, and temporal predictor without oracle information such as the true number of fish.
It retains the training affinity and temporal-cost forms, with an inference association threshold of 0.5 (and a stricter 0.8 acceptance threshold for two-camera groups).

\paragraph{Pass 1: Per-frame encoding and grouping.}\label{app:infer_pass1}

For each frame $t$ in the sequence:
\begin{enumerate}[nosep]
    \item \textbf{Load and encode:} All 2D detections across $C$ cameras are encoded by the frozen encoder~$\phi$, producing embeddings $F \in \mathbb{R}^{N \times 128}$.
    \item \textbf{RoPE and transformer:} Embeddings are projected, augmented with 2D RoPE based on normalised bounding-box centres, and processed by the frozen association transformer~$\psi$, yielding contextualised representations $H \in \mathbb{R}^{N \times 128}$.
    \item \textbf{Pairwise prediction:} For all ordered cross-camera pairs $(i,j)$ with $c_i \neq c_j$, the pairwise MLP outputs directional association probabilities $P_{ij}$.
    \item \textbf{Group and triangulate:} The hybrid affinity matrix $A = 0.6\,P + 0.4\,\frac{\cos(H, H)+1}{2}$ is computed. For each camera pair $(c_a,c_b)$, Hungarian matching uses the directed sub-matrix with rows from camera $c_a$ and columns from camera $c_b$, without symmetrising $P_{ij}$ and $P_{ji}$. Connected components form groups. Each group is pruned to one detection per camera (keeping highest detection score), and multi-view triangulation via DLT yields a 3D position~$\mathbf{X}_g$. Groups whose reprojected 3D point falls outside any member's bounding box are rejected. Overlapping groups (sharing detections) are resolved by keeping the higher-confidence group.
    \item \textbf{Group embedding:} Each group's embedding is the $\ell_2$-normalised mean of its members' transformer outputs:
    $\mathbf{z}_g = \frac{\overline{H}_{\mathrm{members}}}{\|\overline{H}_{\mathrm{members}}\|_2}$.
\end{enumerate}

Pass~1 runs entirely under \texttt{torch.no\_grad()} with optional FP16 mixed precision on GPU.

\paragraph{Pass 2: Temporal tracking.}\label{app:infer_pass2}

Active tracks are maintained in a list, each storing: label, last 3D position, last embedding (both NumPy and Torch), velocity estimate, gap count, and birth/last-seen timestamps.
For each frame:

\paragraph{Track-to-group matching.}
Active tracks are linked to current-frame groups via Hungarian matching with the same weighted distance-and-similarity cost form used during training:
\begin{equation}
    \mathrm{cost}(i,j) = \alpha \cdot \frac{d_{3\mathrm{D}}(\hat{\mathbf{X}}_i, \mathbf{X}_j)}{\theta_{\mathrm{dist}}^{\mathrm{local}}} + \beta \cdot (1 - \cos(\hat{\mathbf{z}}_i, \mathbf{z}_j))
\end{equation}
where $\hat{\mathbf{X}}_i$ and $\hat{\mathbf{z}}_i$ are the \emph{predicted} position and embedding for track~$i$.

\paragraph{PredictorM embedding rollforward.}
When a track has been unobserved for $g > 0$ frames and the trained temporal predictor~$\rho$ is available, the track's embedding is rolled forward by iteratively applying $\rho$ (capped at 5 iterations to limit drift):
$\hat{\mathbf{z}} = \ell_2\text{-norm}(\rho^{(\min(g,5))}(\mathbf{z}_{\mathrm{last}}))$.
This is the direct inference-time payoff of the temporal prediction loss~$\mathcal{L}_{\mathrm{temp}}$.

\paragraph{Constant-velocity position prediction.}
Position is predicted using exponentially smoothed velocity:
$\hat{\mathbf{X}} = \mathbf{X}_{\mathrm{last}} + \mathbf{v} \cdot (g + 1)$,
where $\mathbf{v}$ is updated at each observation as $\mathbf{v} \leftarrow 0.7\,\mathbf{v}_{\mathrm{new}} + 0.3\,\mathbf{v}_{\mathrm{old}}$.

\paragraph{Gate relaxation.}
For tracks with gap $g > 0$, the distance and similarity gates are relaxed using the same formulas as training's fallback matching:
\begin{equation}
    \theta_{\mathrm{dist}}^{\mathrm{local}} = \theta_{\mathrm{dist}} \cdot (1 + 0.15 \cdot g), \quad
    \theta_{\mathrm{sim}}^{\mathrm{local}} = \max(0.1,\; \theta_{\mathrm{sim}} - 0.04 \cdot g)
\end{equation}

\paragraph{Track birth and death.}
Groups that remain unmatched after Hungarian assignment spawn new tracks.
Tracks that have not been observed for more than $\texttt{max\_gap} = 15$ consecutive frames are killed and moved to a graveyard for potential post-hoc merging.

\paragraph{Pass 2.5: Post-hoc track merge.}\label{app:infer_merge}

After temporal tracking, short-lived track fragments may correspond to the same fish observed before and after a long occlusion.
Post-hoc merging identifies such fragments and consolidates them:

\begin{enumerate}[nosep]
    \item All dead tracks (including those still active at the end of the sequence) are sorted chronologically by birth time.
    \item For each fragment~$B$, we search all earlier-ending fragments~$A$ where $A$'s last frame precedes $B$'s first frame (non-overlapping).
    \item A merge is considered if: (a)~the 3D distance between $A$'s last position and $B$'s first position is $\leq 2.0$, and (b)~the cosine similarity between $A$'s last embedding and $B$'s first embedding exceeds $0.45$ (either raw or after rolling $A$'s embedding forward with $\rho$).
    \item The best merge candidate maximises: $\mathrm{score} = \mathrm{sim} - 0.1 \cdot \frac{\mathrm{gap}}{100} - 0.05 \cdot \frac{d_{3\mathrm{D}}}{\theta_{\mathrm{merge}}}$, where $\theta_{\mathrm{merge}} = 2.0$ is the merge distance threshold.
    \item Accepted merges relabel all of $B$'s observations to $A$'s label.
\end{enumerate}

\paragraph{Post-processing.}\label{app:infer_postproc}

Two post-processing steps are applied to all methods (\name and baselines) identically for fairness:

\paragraph{Gap filling.}
For each track, if two consecutive observations are separated by $1 < \Delta t \leq \texttt{max\_gap}$ frames, intermediate frames are filled by linear interpolation of the 3D position.

\paragraph{Ghost filtering.}
Tracks with fewer than $\texttt{min\_obs} = 2$ total observations are removed as likely detection noise or spurious groupings.

After post-processing, tracks are relabelled to consecutive identifiers (A, B, C, \ldots) sorted by first appearance time, and the output is written as a CSV file with columns: \texttt{object, Timestamp, X, Y, Z, group}.

\section{Dataset Statistics}\label{app:dataset_stats}

\begin{table}[H]
\centering
\caption{
Dataset summary and difficulty statistics.
Occlusion level is derived from the OccScore, which is computed per scene by taking, for each (frame, camera) pair, the maximum pairwise 2D bounding-box IoU among all tracked objects, then averaging these maxima across all frames and camera views.
\%Overlap is the fraction of (frame, camera) instances with any overlap (IoU$>$0.01).
}
\label{tab:dataset_summary}
\resizebox{0.85\textwidth}{!}{
\begin{tabular}{llcccccc}
\hline
\textbf{Split} & \textbf{Scene} & \textbf{\#Fish} & \textbf{\#Frames} & \textbf{\#Cams} & \textbf{Resolution} & \textbf{OcclLevel} & \textbf{\%Overlap} \\
\hline
SynFish (train) & train1   & 4  & 359 & 3 & 1920$\times$1080 & Low      & 0.2\% \\
SynFish (train) & train2   & 6  & 349 & 3 & 1920$\times$1080 & Mild     & 10.2\% \\
SynFish (train) & train3   & 8  & 362 & 3 & 1920$\times$1080 & Moderate & 81.1\% \\
SynFish (train) & train4   & 10 & 359 & 3 & 1920$\times$1080 & Moderate & 75.1\% \\
SynFish (train) & train5   & 14 & 359 & 3 & 1920$\times$1080 & Heavy    & 91.9\% \\
SynFish (train) & train6   & 14 & 362 & 3 & 1920$\times$1080 & Mild     & 66.9\% \\
SynFish (train) & train7   & 16 & 362 & 3 & 1920$\times$1080 & Mild     & 74.1\% \\
SynFish (train) & train8   & 16 & 362 & 3 & 1920$\times$1080 & Moderate & 99.7\% \\
\hline
SynFish (test)  & test1    & 4  & 359 & 3 & 1920$\times$1080 & Low      & 0.2\% \\
SynFish (test)  & test2    & 5  & 359 & 3 & 1920$\times$1080 & Low      & 0.3\% \\
SynFish (test)  & test3    & 8  & 359 & 3 & 1920$\times$1080 & Moderate & 68.0\% \\
SynFish (test)  & test4    & 16 & 362 & 3 & 1920$\times$1080 & Heavy    & 100.0\% \\
SynFish (test)  & test5    & 16 & 362 & 3 & 1920$\times$1080 & Heavy    & 99.9\% \\
SynFish (test)  & test6    & 16 & 362 & 3 & 1920$\times$1080 & Mild     & 99.1\% \\
\hline
3D-ZeF (train) & zebra01 & 2 & 800 & 2 & 2704$\times$1520 & Moderate & 69.4\% \\
3D-ZeF (train) & zebra03 & 2 & 800 & 2 & 2704$\times$1520 & Low      & 16.8\% \\
\hline
3D-ZeF (test)  & zebra02 & 5 & 701 & 2 & 2704$\times$1520 & Moderate & 83.8\% \\
3D-ZeF (test)  & zebra04 & 5 & 910 & 2 & 2704$\times$1520 & Heavy    & 86.8\% \\
\hline
\end{tabular}
}
\end{table}

\section{Camera Calibration Parameters}\label{app:camera_params}

This section lists the complete intrinsic and extrinsic camera calibration for every dataset scene.
Extrinsics are reported as the rotation matrix $R \in \mathrm{SO}(3)$ and translation vector $\mathbf{t} \in \mathbb{R}^3$ such that a world point $\mathbf{X}$ projects to pixel $\mathbf{u} = K[R|\mathbf{t}]\mathbf{X}$.

\paragraph{SynFish Synthetic Scenes.}\label{app:cam_ue5}

All 14 SynFish scenes are rendered at $1920 \times 1080$ with $C{=}3$ synchronised pinhole cameras sharing identical intrinsics:
\begin{equation}
    K_{\mathrm{syn}} = \begin{bmatrix} 1866.67 & 0 & 960.0 \\ 0 & 1866.67 & 540.0 \\ 0 & 0 & 1 \end{bmatrix}
    \quad (\text{HFoV} \approx 54.3^{\circ})
\end{equation}

\noindent Camera extrinsics are defined by \textbf{7 distinct rig configurations} (numbered 0--6).
Scenes sharing the same rig use identical camera placements.
The training set spans 6 rig configurations; the test set uses 4, of which one (rig~0) is unseen during training.
Table~\ref{tab:scene_rig} maps each scene to its rig, and Table~\ref{tab:ue5_extrinsics} lists the full extrinsic parameters.

\begin{table}[H]
\centering
\caption{SynFish scene-to-rig mapping. Scenes sharing a rig use identical camera placements (same $R$ and $\mathbf{t}$ for all 3 cameras). Rig~0 appears only in the test set.}
\label{tab:scene_rig}
\small
\begin{tabular}{lll}
\hline
\textbf{Rig} & \textbf{Train scenes} & \textbf{Test scenes} \\
\hline
0 & --- & test1, test2 \\
1 & train1 & --- \\
2 & train2 & --- \\
3 & train4 & --- \\
4 & train5, train6 & --- \\
5 & train3 & test3 \\
6 & train7, train8 & test4, test5, test6 \\
\hline
\end{tabular}
\end{table}

\begin{table}[H]
\centering
\caption{SynFish camera rig extrinsics. Each rig defines 3 cameras (0, 1, 2) with rotation $R$ (shown row-by-row) and translation $\mathbf{t}$. All cameras share the intrinsic $K_{\mathrm{syn}}$ above.}
\label{tab:ue5_extrinsics}
\scriptsize
\setlength{\tabcolsep}{4pt}
\begin{tabular}{ll rrr rrr}
\hline
\textbf{Rig} & \textbf{Cam} & \multicolumn{3}{c}{$R$ (row-major)} & \multicolumn{3}{c}{$\mathbf{t}$} \\
\hline
\texttt{0} & 0 &   $-$0.6129 &    0.7902 &   $-$0.0000 &     7.48 &    18.22 &    43.67 \\
 &  &    0.1589 &    0.1232 &   $-$0.9796 & & & \\
 &  &   $-$0.7740 &   $-$0.6004 &   $-$0.2011 & & & \\
 & 1 &   $-$0.9866 &   $-$0.1633 &    0.0000 &   $-$1.35 &    13.66 &    73.22 \\
 &  &   $-$0.1450 &    0.8759 &   $-$0.4602 & & & \\
 &  &    0.0752 &   $-$0.4540 &   $-$0.8878 & & & \\
 & 2 &   $-$0.9282 &    0.3721 &   $-$0.0000 &     5.41 &    24.25 &    34.57 \\
 &  &   $-$0.0298 &   $-$0.0744 &   $-$0.9968 & & & \\
 &  &   $-$0.3709 &   $-$0.9252 &    0.0802 & & & \\
\hline
\texttt{1} & 0 &   $-$0.6129 &    0.7902 &   $-$0.0000 &     8.20 &    19.80 &    43.94 \\
 &  &    0.1589 &    0.1232 &   $-$0.9796 & & & \\
 &  &   $-$0.7740 &   $-$0.6004 &   $-$0.2011 & & & \\
 & 1 &   $-$0.9866 &   $-$0.1633 &    0.0000 &   $-$0.54 &    13.67 &    66.57 \\
 &  &   $-$0.1450 &    0.8759 &   $-$0.4602 & & & \\
 &  &    0.0752 &   $-$0.4540 &   $-$0.8878 & & & \\
 & 2 &   $-$0.9282 &    0.3721 &   $-$0.0000 &    11.51 &    23.96 &    42.23 \\
 &  &   $-$0.0298 &   $-$0.0744 &   $-$0.9968 & & & \\
 &  &   $-$0.3709 &   $-$0.9252 &    0.0802 & & & \\
\hline
\texttt{2} & 0 &   $-$0.7547 &    0.6561 &    0.0000 &    13.00 &    18.33 &    29.30 \\
 &  &    0.0936 &    0.1076 &   $-$0.9898 & & & \\
 &  &   $-$0.6494 &   $-$0.7470 &   $-$0.1426 & & & \\
 & 1 &   $-$0.9511 &    0.3090 &   $-$0.0000 &     4.26 &    26.58 &    59.48 \\
 &  &    0.2654 &    0.8169 &   $-$0.5120 & & & \\
 &  &   $-$0.1582 &   $-$0.4870 &   $-$0.8590 & & & \\
 & 2 &    0.9511 &   $-$0.3090 &    0.0000 &   $-$6.22 &    24.97 &    51.34 \\
 &  &    0.1026 &    0.3159 &   $-$0.9432 & & & \\
 &  &    0.2915 &    0.8971 &    0.3322 & & & \\
\hline
\texttt{3} & 0 &   $-$0.6129 &    0.7902 &   $-$0.0000 &     7.48 &    18.22 &    43.67 \\
 &  &    0.1589 &    0.1232 &   $-$0.9796 & & & \\
 &  &   $-$0.7740 &   $-$0.6004 &   $-$0.2011 & & & \\
 & 1 &   $-$0.9866 &   $-$0.1633 &    0.0000 &   $-$1.35 &    13.66 &    73.22 \\
 &  &   $-$0.1450 &    0.8759 &   $-$0.4602 & & & \\
 &  &    0.0752 &   $-$0.4540 &   $-$0.8878 & & & \\
 & 2 &   $-$0.9311 &    0.3649 &   $-$0.0000 &     5.70 &    24.25 &    34.52 \\
 &  &   $-$0.0293 &   $-$0.0747 &   $-$0.9968 & & & \\
 &  &   $-$0.3637 &   $-$0.9281 &    0.0802 & & & \\
\hline
\texttt{4} & 0 &    0.9499 &    0.3126 &   $-$0.0000 &   $-$2.72 &    10.39 &    72.20 \\
 &  &    0.1146 &   $-$0.3482 &   $-$0.9304 & & & \\
 &  &   $-$0.2908 &    0.8838 &   $-$0.3666 & & & \\
 & 1 &   $-$0.9866 &   $-$0.1633 &    0.0000 &   $-$0.36 &    15.77 &    70.33 \\
 &  &   $-$0.1450 &    0.8759 &   $-$0.4602 & & & \\
 &  &    0.0752 &   $-$0.4540 &   $-$0.8878 & & & \\
 & 2 &   $-$0.9282 &    0.3721 &   $-$0.0000 &     5.98 &    19.98 &    45.08 \\
 &  &   $-$0.0298 &   $-$0.0744 &   $-$0.9968 & & & \\
 &  &   $-$0.3709 &   $-$0.9252 &    0.0802 & & & \\
\hline
\texttt{5} & 0 &   $-$0.6129 &    0.7902 &   $-$0.0000 &     4.87 &    17.82 &    45.65 \\
 &  &    0.1589 &    0.1232 &   $-$0.9796 & & & \\
 &  &   $-$0.7740 &   $-$0.6004 &   $-$0.2011 & & & \\
 & 1 &   $-$0.9866 &   $-$0.1633 &    0.0000 &   $-$1.42 &    13.00 &    71.09 \\
 &  &   $-$0.1450 &    0.8759 &   $-$0.4602 & & & \\
 &  &    0.0752 &   $-$0.4540 &   $-$0.8878 & & & \\
 & 2 &   $-$0.5210 &   $-$0.8536 &    0.0000 &   $-$1.94 &    22.44 &    31.22 \\
 &  &    0.0685 &   $-$0.0418 &   $-$0.9968 & & & \\
 &  &    0.8508 &   $-$0.5193 &    0.0802 & & & \\
\hline
\texttt{6} & 0 &    0.9499 &    0.3126 &   $-$0.0000 &   $-$0.59 &    13.00 &    68.85 \\
 &  &    0.1146 &   $-$0.3482 &   $-$0.9304 & & & \\
 &  &   $-$0.2908 &    0.8838 &   $-$0.3666 & & & \\
 & 1 &   $-$0.9866 &   $-$0.1633 &    0.0000 &   $-$0.54 &    16.15 &    70.02 \\
 &  &   $-$0.1450 &    0.8759 &   $-$0.4602 & & & \\
 &  &    0.0752 &   $-$0.4540 &   $-$0.8878 & & & \\
 & 2 &   $-$0.9311 &    0.3649 &   $-$0.0000 &     6.34 &    17.98 &    45.20 \\
 &  &   $-$0.0293 &   $-$0.0747 &   $-$0.9968 & & & \\
 &  &   $-$0.3637 &   $-$0.9281 &    0.0802 & & & \\
\hline
\end{tabular}
\end{table}

\paragraph{3D-ZeF Zebrafish Sequences.}\label{app:cam_zef}

The 3D-ZeF benchmark~\cite{pedersen2020zef} uses $C{=}2$ cameras at $2704 \times 1520$ resolution.
Both cameras retain \emph{fixed intrinsics} across all four sequences (the physical cameras are not changed between recordings):
\begin{equation}
    K_0 = \begin{bmatrix} 1490.78 & 0 & 1343.87 \\ 0 & 1463.64 & 781.86 \\ 0 & 0 & 1 \end{bmatrix}, \quad
    K_1 = \begin{bmatrix} 1460.43 & 0 & 1348.20 \\ 0 & 1457.88 & 774.27 \\ 0 & 0 & 1 \end{bmatrix}
\end{equation}
Note that unlike the SynFish cameras, $K_0$ and $K_1$ have different focal lengths and principal points because they are two distinct physical cameras.
The extrinsics vary per sequence because the cameras are repositioned slightly between recording sessions (Table~\ref{tab:zef_extrinsics}).

\begin{table}[H]
\centering
\caption{3D-ZeF camera extrinsics per sequence. All sequences share the intrinsics $K_0$, $K_1$ above.}
\label{tab:zef_extrinsics}
\scriptsize
\setlength{\tabcolsep}{4pt}
\begin{tabular}{ll rrr rrr}
\hline
\textbf{Sequence} & \textbf{Cam} & \multicolumn{3}{c}{$R$ (row-major)} & \multicolumn{3}{c}{$\mathbf{t}$} \\
\hline
zebra01 (train) & 0 &    0.9999 &   $-$0.0143 &    0.0054 &   $-$14.68 &   $-$16.34 &    31.24 \\
 &  &    0.0139 &    0.9974 &    0.0708 & & & \\
 &  &   $-$0.0064 &   $-$0.0707 &    0.9975 & & & \\
 & 1 &    0.9999 &   $-$0.0077 &   $-$0.0067 &   $-$14.63 &    $-$9.80 &    48.50 \\
 &  &    0.0079 &    0.1591 &    0.9872 & & & \\
 &  &   $-$0.0065 &   $-$0.9872 &    0.1591 & & & \\
\hline
zebra03 (train) & 0 &    0.9989 &   $-$0.0109 &    0.0453 &   $-$15.03 &   $-$15.83 &    33.80 \\
 &  &    0.0076 &    0.9973 &    0.0726 & & & \\
 &  &   $-$0.0460 &   $-$0.0722 &    0.9963 & & & \\
 & 1 &    0.9987 &   $-$0.0487 &   $-$0.0128 &   $-$13.18 &   $-$10.62 &    48.76 \\
 &  &    0.0216 &    0.1857 &    0.9824 & & & \\
 &  &   $-$0.0455 &   $-$0.9814 &    0.1866 & & & \\
\hline
zebra02 (test) & 0 &    0.9990 &   $-$0.0115 &    0.0430 &   $-$14.97 &   $-$15.95 &    32.52 \\
 &  &    0.0092 &    0.9986 &    0.0528 & & & \\
 &  &   $-$0.0435 &   $-$0.0524 &    0.9977 & & & \\
 & 1 &    0.9994 &   $-$0.0332 &   $-$0.0100 &   $-$13.73 &   $-$10.09 &    48.67 \\
 &  &    0.0154 &    0.1669 &    0.9859 & & & \\
 &  &   $-$0.0311 &   $-$0.9854 &    0.1673 & & & \\
\hline
zebra04 (test) & 0 &    0.9999 &   $-$0.0159 &    0.0011 &   $-$15.09 &   $-$14.85 &    33.05 \\
 &  &    0.0158 &    0.9981 &    0.0594 & & & \\
 &  &   $-$0.0020 &   $-$0.0594 &    0.9982 & & & \\
 & 1 &    0.9988 &    0.0458 &   $-$0.0148 &   $-$15.74 &   $-$11.04 &    46.73 \\
 &  &    0.0048 &    0.2121 &    0.9772 & & & \\
 &  &    0.0479 &   $-$0.9762 &    0.2117 & & & \\
\hline
\end{tabular}
\end{table}

\section{Per-Scene Results: SynFish Test Set}\label{app:ue5_perscene}

\begin{table}[H]
\centering
\caption{Per-scene accuracy metrics on the SynFish test set (6 held-out scenes).
Scene names correspond to Table~\ref{tab:dataset_summary}: test1 (4 fish), test2 (5 fish), test3 (8 fish), test4--test6 (16 fish each).
}
\label{tab:app_ue5_perscene}
\resizebox{\textwidth}{!}{
\scriptsize
\begin{tabular}{l rrrrrrr rrrrrrr rrrrrrr}
\hline
& \multicolumn{7}{c}{\textbf{MOTA (\%) $\uparrow$}} & \multicolumn{7}{c}{\textbf{MT (\%) $\uparrow$}} & \multicolumn{7}{c}{\textbf{ML (\%) $\downarrow$}} \\
\textbf{Method} & \textbf{test1} & \textbf{test2} & \textbf{test3} & \textbf{test4} & \textbf{test5} & \textbf{test6} & \textbf{Avg}
& \textbf{test1} & \textbf{test2} & \textbf{test3} & \textbf{test4} & \textbf{test5} & \textbf{test6} & \textbf{Avg}
& \textbf{test1} & \textbf{test2} & \textbf{test3} & \textbf{test4} & \textbf{test5} & \textbf{test6} & \textbf{Avg} \\
\hline
\name (SAM3)   & 99.6 & 99.9 & 96.2 & 86.2 & 94.8 & 98.0 & 95.8  & 100 & 100 & 100 & 81.2 & 93.8 & 100 & 95.8  & 0 & 0 & 0 & 0 & 0 & 0 & 0.0 \\
\name (YOLO26+SAM2) & 100 & 100 & 97.7 & 86.3 & 96.7 & 99.1 & 96.6  & 100 & 100 & 100 & 81.2 & 93.8 & 100 & 95.8  & 0 & 0 & 0 & 0 & 0 & 0 & 0.0 \\
SAM3+Hung.         & 100 & 100 & 93.8 & 55.2 & 75.1 & 77.1 & 83.5  & 100 & 100 & 100 & 31.2 & 43.8 & 56.2 & 71.9  & 0 & 0 & 0 & 0 & 0 & 0 & 0.0 \\
SAM3+Greedy        & 100 & 100 & 93.8 & 55.2 & 75.1 & 77.1 & 83.5  & 100 & 100 & 100 & 31.2 & 43.8 & 56.2 & 71.9  & 0 & 0 & 0 & 0 & 0 & 0 & 0.0 \\
YOLO26+SAM2+Hung.  & 99.7 & 99.8 & 95.9 & 53.7 & 69.0 & 75.0 & 82.2  & 100 & 100 & 100 & 25.0 & 43.8 & 50.0 & 69.8  & 0 & 0 & 0 & 0 & 0 & 0 & 0.0 \\
YOLO26+SAM2+Greedy & 99.7 & 99.8 & 95.9 & 53.7 & 69.0 & 75.0 & 82.2  & 100 & 100 & 100 & 25.0 & 43.8 & 50.0 & 69.8  & 0 & 0 & 0 & 0 & 0 & 0 & 0.0 \\
Self-MVA           & $-$65.0 & $-$75.9 & $-$44.5 & $-$72.4 & $-$66.3 & $-$63.4 & $-$64.6  & 0 & 0 & 0 & 0 & 0 & 0 & 0.0  & 100 & 100 & 87.5 & 100 & 100 & 93.8 & 96.9 \\
ASNet              & 82.1 & 60.8 & 58.5 & $-$0.8 & 6.1 & $-$6.9 & 33.3  & 100 & 60 & 75 & 12.5 & 31.2 & 18.8 & 49.6  & 0 & 0 & 12.5 & 12.5 & 18.8 & 25 & 11.5 \\
SambaMOTR          & $-$0.3 & $-$26.8 & 16.8 & $-$14.3 & $-$19.8 & $-$18.8 & $-$10.5  & 25 & 20 & 12.5 & 6.2 & 6.2 & 0 & 11.7  & 25 & 80 & 50 & 93.8 & 93.8 & 100 & 73.8 \\
MOTIP              & 44.0 & 48.0 & $-$1.4 & 3.9 & $-$0.2 & $-$6.1 & 14.7  & 50 & 60 & 0 & 6.2 & 6.2 & 6.2 & 21.4  & 25 & 20 & 25 & 68.8 & 75 & 43.8 & 42.9 \\
ReST               & 81.0 & 73.1 & 80.4 & 54.2 & 32.4 & 46.8 & 61.3  & 100 & 80 & 100 & 62.5 & 43.8 & 62.5 & 74.8  & 0 & 0 & 0 & 0 & 0 & 0 & 0.0 \\
MCTR               & 99.4 & 46.9 & 86.9 & 83.3 & 88.0 & 97.5 & 83.7  & 100 & 40 & 87.5 & 81.2 & 87.5 & 100 & 82.7  & 0 & 0 & 0 & 0 & 0 & 0 & 0.0 \\
\hline
\end{tabular}
}
\end{table}

\begin{table}[H]
\centering
\caption{Per-scene identity stability metrics on the SynFish test set (6 held-out scenes). Same scene abbreviations as Table~\ref{tab:app_ue5_perscene}.}
\label{tab:app_ue5_perscene_id}
\resizebox{\textwidth}{!}{
\scriptsize
\begin{tabular}{l rrrrrrr rrrrrrr rrrrrrr}
\hline
& \multicolumn{7}{c}{\textbf{IDSw $\downarrow$}} & \multicolumn{7}{c}{\textbf{Frag $\downarrow$}} & \multicolumn{7}{c}{\textbf{MTBF $\uparrow$}} \\
\textbf{Method} & \textbf{test1} & \textbf{test2} & \textbf{test3} & \textbf{test4} & \textbf{test5} & \textbf{test6} & \textbf{Avg}
& \textbf{test1} & \textbf{test2} & \textbf{test3} & \textbf{test4} & \textbf{test5} & \textbf{test6} & \textbf{Avg}
& \textbf{test1} & \textbf{test2} & \textbf{test3} & \textbf{test4} & \textbf{test5} & \textbf{test6} & \textbf{Avg} \\
\hline
\name (SAM3)   & 0 & 0 & 0 & 0 & 0 & 0 & 0.0  & 0 & 0 & 2 & 18 & 9 & 10 & 6.5  & 359.0 & 358.6 & 276.4 & 158.6 & 224.7 & 220.5 & 266.3 \\
\name (YOLO26+SAM2) & 0 & 0 & 0 & 0 & 0 & 1 & 0.2  & 0 & 0 & 0 & 11 & 5 & 6 & 3.7  & 359.0 & 359.0 & 350.6 & 198.5 & 271.2 & 240.5 & 296.5 \\
SAM3+Hung.         & 0 & 0 & 5 & 12 & 17 & 11 & 7.5  & 0 & 0 & 6 & 27 & 25 & 19 & 12.8  & 359.0 & 359.0 & 142.4 & 66.0 & 77.8 & 98.3 & 183.8 \\
SAM3+Greedy        & 0 & 0 & 5 & 12 & 17 & 11 & 7.5  & 0 & 0 & 6 & 27 & 25 & 19 & 12.8  & 359.0 & 359.0 & 142.4 & 66.0 & 77.8 & 98.3 & 183.8 \\
YOLO26+SAM2+Hung.  & 0 & 0 & 3 & 21 & 13 & 15 & 8.7  & 0 & 0 & 7 & 27 & 18 & 21 & 12.2  & 358.0 & 358.4 & 154.1 & 54.1 & 88.2 & 85.9 & 183.1 \\
YOLO26+SAM2+Greedy & 0 & 0 & 3 & 21 & 13 & 15 & 8.7  & 0 & 0 & 7 & 27 & 18 & 21 & 12.2  & 358.0 & 358.4 & 154.1 & 54.1 & 88.2 & 85.9 & 183.1 \\
Self-MVA           & 0 & 0 & 6 & 2 & 1 & 1 & 1.7  & 0 & 0 & 6 & 1 & 3 & 3 & 2.2  & 34.0 & 9.0 & 16.6 & 10.6 & 10.1 & 14.9 & 15.9 \\
ASNet              & 2 & 5 & 8 & 34 & 37 & 39 & 20.8  & 5 & 5 & 17 & 44 & 47 & 56 & 29.0  & 116.1 & 92.9 & 69.0 & 32.3 & 30.4 & 23.5 & 60.7 \\
SambaMOTR          & 0 & 0 & 0 & 0 & 0 & 0 & 0.0  & 26 & 3 & 9 & 2 & 7 & 0 & 7.8  & 24.7 & 74.5 & 73.8 & 103.0 & 34.8 & 0.0 & 51.8 \\
MOTIP              & 0 & 0 & 0 & 0 & 0 & 0 & 0.0  & 30 & 19 & 29 & 18 & 18 & 40 & 25.7  & 30.4 & 55.8 & 34.9 & 45.9 & 44.9 & 35.3 & 41.2 \\
ReST               & 12 & 51 & 77 & 228 & 334 & 340 & 173.7  & 5 & 23 & 26 & 114 & 174 & 129 & 78.5  & 49.9 & 15.8 & 16.6 & 9.8 & 6.2 & 6.8 & 17.5 \\
MCTR               & 0 & 7 & 3 & 0 & 0 & 0 & 1.7  & 1 & 14 & 17 & 32 & 10 & 11 & 14.2  & 286.4 & 48.5 & 96.9 & 110.6 & 209.1 & 211.8 & 160.6 \\
\hline
\end{tabular}
}
\end{table}

\section{Per-Scene Results: 3D-ZeF Test Set}\label{app:zebra_perscene}

\begin{table}[H]
\centering
\caption{Per-scene 3D tracking results on the 3D-ZeF test set (zebra2: 5 fish, zebra4: 5 fish).
ReST fails to produce any valid cross-camera associations on zebrafish sequences.
}
\label{tab:app_zebra_perscene}
\resizebox{\textwidth}{!}{
\scriptsize
\begin{tabular}{l rrr rrr rrr rrr rrr rrr}
\hline
& \multicolumn{3}{c}{\textbf{MOTA\% $\uparrow$}} & \multicolumn{3}{c}{\textbf{MT\% $\uparrow$}} & \multicolumn{3}{c}{\textbf{ML\% $\downarrow$}} & \multicolumn{3}{c}{\textbf{IDSw $\downarrow$}} & \multicolumn{3}{c}{\textbf{Frag $\downarrow$}} & \multicolumn{3}{c}{\textbf{MTBF $\uparrow$}} \\
\textbf{Method} & \textbf{z2} & \textbf{z4} & \textbf{Avg} & \textbf{z2} & \textbf{z4} & \textbf{Avg} & \textbf{z2} & \textbf{z4} & \textbf{Avg} & \textbf{z2} & \textbf{z4} & \textbf{Avg} & \textbf{z2} & \textbf{z4} & \textbf{Avg} & \textbf{z2} & \textbf{z4} & \textbf{Avg} \\
\hline
\name   & 82.7 & 79.4 & 81.1  & 100 & 80 & 90.0  & 0 & 0 & 0.0  & 2 & 6 & 4.0  & 24 & 37 & 30.5  & 103.9 & 83.3 & 93.6 \\
Hung.+NN    & 77.1 & 77.7 & 77.4  & 100 & 80 & 90.0  & 0 & 0 & 0.0  & 6 & 6 & 6.0  & 60 & 86 & 73.0  & 43.8 & 41.2 & 42.5 \\
Greedy+NN   & 76.3 & 77.4 & 76.9  & 100 & 80 & 90.0  & 0 & 0 & 0.0  & 8 & 12 & 10.0  & 64 & 88 & 76.0  & 40.3 & 37.1 & 38.7 \\
Self-MVA    & 84.7 & 16.8 & 50.8  & 100 & 20 & 60.0  & 0 & 0 & 0.0  & 2 & 41 & 21.5  & 33 & 114 & 73.5  & 81.2 & 15.8 & 48.5 \\
ASNet       & 13.9 & $-$13.4 & 0.3  & 0 & 0 & 0.0  & 0 & 0 & 0.0  & 55 & 90 & 72.5  & 152 & 198 & 175.0  & 11.5 & 9.7 & 10.6 \\
SambaMOTR   & $-$31.3 & $-$21.4 & $-$26.4  & 0 & 0 & 0.0  & 100 & 100 & 100.0  & 0 & 1 & 0.5  & 0 & 2 & 1.0  & 7.0 & 6.0 & 6.5 \\
MOTIP       & $-$34.1 & $-$57.6 & $-$45.9  & 0 & 0 & 0.0  & 100 & 100 & 100.0  & 1 & 2 & 1.5  & 4 & 9 & 6.5  & 3.3 & 2.1 & 2.7 \\
ReST        & 0.0 & 0.0 & 0.0  & 0 & 0 & 0.0  & 100 & 100 & 100.0  & 0 & 0 & 0.0  & 0 & 0 & 0.0  & 0.0 & 0.0 & 0.0 \\
MCTR        & $-$1.2 & 1.3 & 0.1  & 0 & 0 & 0.0  & 0 & 80 & 40.0  & 27 & 15 & 21.0  & 68 & 47 & 57.5  & 8.7 & 8.8 & 8.8 \\
\hline
\end{tabular}
}
\end{table}

\section{Temporal Linking Weight Ablation}\label{app:temporal_weight_ablation}

During training, the temporal linking weights $(\alpha,\beta)$ govern which cross-frame group pairs are selected as positive examples for the temporal prediction loss $\mathcal{L}_\mathrm{temp}$ (Sec.~3.5).
We set $\alpha{=}0.6,\,\beta{=}0.4$ so that matched pairs must satisfy both spatial proximity \emph{and} embedding agreement, yielding a high-quality supervision signal for the predictor~$\mathcal{M}$: pairs that are merely close in 3D but visually dissimilar are down-weighted in the cost matrix, reducing noisy gradients.

At inference, the same $(\alpha,\beta)$ parameterises the temporal linking cost (Sec.~3.6).
Table~\ref{tab:temporal_weight_ablation} ablates this weighting on the SynFish test set (6 held-out scenes) with all other components fixed.
Performance is stable across a wide operating range ($\alpha \in [0.4, 1.0]$), confirming that the upstream cross-camera association produces well-separated 3D trajectories where spatial proximity is already a reliable temporal cue.
Degradation appears only at the embedding similarity-only extreme ($\alpha{=}0,\,\beta{=}1$), where removing the 3D distance signal increases identity switches and fragmentation.
We therefore adopt $\alpha{=}0.6,\,\beta{=}0.4$ as it (i)~matches the training regime, ensuring consistency between the distribution of temporal matches seen during training and those encountered at inference, and (ii)~retains the learned embedding similarity as a complementary cue for disambiguating spatially proximate individuals in denser scenarios.

\begin{table}[H]
\centering
\caption{Ablation of temporal linking weights $(\alpha,\beta)$ on the SynFish test set (6 scenes, averaged).
The cost is robust for $\alpha \geq 0.4$; removing the distance component ($\alpha{=}0$) degrades identity stability.
The selected setting is \textbf{bolded}.}
\label{tab:temporal_weight_ablation}
\small
\begin{tabular}{cc rrrrrr}
\hline
$\alpha$ & $\beta$ & \textbf{MOTA\,$\uparrow$} & \textbf{MT\%\,$\uparrow$} & \textbf{ML\%\,$\downarrow$} & \textbf{IDSw\,$\downarrow$} & \textbf{Frag\,$\downarrow$} & \textbf{MTBF\,$\uparrow$} \\
\hline
1.0 & 0.0 & 95.8\% & 95.8\% & 0.0\% & 0.0 & 6.5 & 266.3 \\
0.8 & 0.2 & 95.8\% & 95.8\% & 0.0\% & 0.0 & 6.5 & 266.3 \\
\textbf{0.6} & \textbf{0.4} & \textbf{95.8\%} & \textbf{95.8\%} & \textbf{0.0\%} & \textbf{0.0} & \textbf{6.5} & \textbf{266.3} \\
0.4 & 0.6 & 95.8\% & 95.8\% & 0.0\% & 0.0 & 6.5 & 266.3 \\
0.2 & 0.8 & 95.8\% & 95.8\% & 0.0\% & 0.2 & 6.8 & 262.5 \\
0.0 & 1.0 & 95.8\% & 95.8\% & 0.0\% & 0.3 & 7.5 & 257.3 \\
\hline
\end{tabular}
\end{table}

\section{Robustness to Camera Calibration Noise}\label{app:robustness}

\name relies on calibrated camera extrinsics for geometric pseudo-labelling, triangulation, and temporal linking.
To characterise how sensitive the system is to calibration error, we inject increasing levels of Gaussian noise into the camera extrinsics at \emph{inference time} (the model is trained with the original, unperturbed calibration) and re-evaluate on the SynFish test set.

\paragraph{Perturbation protocol.}
For each noise level~$\sigma$, we independently perturb (i)~rotation by sampling $\delta\boldsymbol{\omega} \sim \mathcal{N}(\mathbf{0},\sigma_R^2 I_3)$ and composing $R' = \exp([\delta\boldsymbol{\omega}]_\times)\,R$ via the Rodrigues formula, and (ii)~translation by adding $\delta\mathbf{t} \sim \mathcal{N}(\mathbf{0},\sigma_t^2 I_3)$.
We set $\sigma_R = \sigma_t = \sigma$ and report results for four noise levels alongside the unperturbed baseline (Table~\ref{tab:robustness}).
All other hyperparameters remain unchanged.

\begin{table}[H]
\centering
\caption{Robustness of \name (SAM\,3) to camera extrinsic perturbation on the SynFish test set (6 scenes, averaged).
$\sigma_R$ is the rotation noise standard deviation (degrees) and $\sigma_t$ is the translation noise standard deviation (world units).
The unperturbed baseline ($\sigma{=}0$) is shown in the first row.
}
\label{tab:robustness}
\small
\begin{tabular}{cc rrrrrr}
\hline
$\sigma_R$ (\textdegree) & $\sigma_t$ & \textbf{MOTA\,$\uparrow$} & \textbf{MT\%\,$\uparrow$} & \textbf{ML\%\,$\downarrow$} & \textbf{IDSw\,$\downarrow$} & \textbf{Frag\,$\downarrow$} & \textbf{MTBF\,$\uparrow$} \\
\hline
0    & 0    & 95.8\%   & 95.8\% & 0.0\%  & 0.0 &  6.5 & 266.3 \\
0.05 & 0.05 & 94.1\%   & 95.8\% & 0.0\%  & 0.0 &  5.0 & 274.0 \\
0.1  & 0.1  & 92.1\%   & 93.8\% & 0.0\%  & 0.0 &  9.2 & 246.6 \\
0.2  & 0.2  & 81.3\%   & 82.3\% & 1.0\%  & 0.2 & 20.3 & 206.0 \\
\hline
\end{tabular}
\end{table}

\paragraph{Discussion.}
The system is robust to small calibration errors: at $\sigma{=}0.05$ the MOTA drops by only 1.7 percentage points (pp) and all tracks remain mostly-tracked.
At $\sigma{=}0.1$ the degradation is still moderate ($-$3.7\,pp MOTA) with no increase in identity switches, though fragment count rises from 6.5 to 9.2 as noisy triangulation occasionally pushes 3D positions outside the matching gate.
Performance degrades more steeply at $\sigma{=}0.2$ ($-$14.5\,pp MOTA, MT\% drops to 82.3\%) because the corrupted projection matrices $P{=}K[R'|t']$ cause the reprojection-based group validation to reject an increasing number of valid cross-camera associations, which in turn produces fewer and less accurate 3D detections.

These results are consistent with the geometry of the problem.
\name solves associations in 2D and lifts them to 3D via Direct Linear Transform (DLT) triangulation; small extrinsic errors propagate quadratically through the $P{=}K[R|t]$ projection and are amplified at greater object distances.
The practical implication is that \name requires calibration accuracy on the order of $\sigma_R \lesssim 0.1$\textdegree\ and $\sigma_t \lesssim 0.1$ world units to maintain near-baseline performance.

\section{Robustness to Detection Noise}\label{app:det_robustness}

In addition to camera calibration errors (Appendix~\ref{app:robustness}), real-world deployments also face imperfect 2D detectors.
We note that \emph{all} main results already operate on noisy detections: the 2D inputs are produced by off-the-shelf models (SAM\,3 or YOLO26+SAM2) finetuned on training data, not from ground-truth bounding boxes.
These detectors inherently introduce false positives (spurious boxes), false negatives (missed fish), and centroid localisation error.
The ablation below therefore injects \emph{additional} synthetic noise on top of this realistic detection noise to stress-test \name's tolerance beyond normal operating conditions.

\paragraph{Perturbation protocol.}
A single noise parameter~$\sigma$ controls both failure modes simultaneously:
\begin{itemize}[nosep]
    \item \textbf{False negatives (FN):} each real detection is independently dropped with probability~$\sigma$.
    \item \textbf{False positives (FP):} for each frame-camera with $n$ real detections, $\lfloor \sigma \cdot n \rfloor$ (plus a probabilistic extra for the fractional part) spurious detections are injected at random image locations with bounding-box sizes sampled from the empirical distribution of real detections and scores drawn from a Beta distribution ($\mathrm{mean} \approx 0.85$) to mimic marginal false positives.
\end{itemize}
We evaluate at four noise levels ($\sigma \in \{0.05, 0.10, 0.20, 0.40\}$) alongside the unperturbed baseline on the SynFish test set (6 scenes).
All experiments use a fixed random seed for reproducibility.

\begin{table}[H]
\centering
\caption{Robustness of \name (SAM\,3) to detection noise on the SynFish test set (6 scenes, averaged).
$\sigma$ is the noise level controlling both the false-negative drop rate and the false-positive injection rate.
The unperturbed baseline ($\sigma{=}0$) is shown in the first row.
}
\label{tab:det_robustness}
\small
\begin{tabular}{c rrrrrr}
\hline
$\sigma$ & \textbf{MOTA\,$\uparrow$} & \textbf{MT\%\,$\uparrow$} & \textbf{ML\%\,$\downarrow$} & \textbf{IDSw\,$\downarrow$} & \textbf{Frag\,$\downarrow$} & \textbf{MTBF\,$\uparrow$} \\
\hline
0    & 95.8\%  & 95.8\% & 0.0\%  & 0.0  &  6.5 & 266.3 \\
0.05 & 90.3\%  & 95.8\% & 0.0\%  & 3.8  & 12.2 & 175.2 \\
0.10 & 84.1\%  & 94.8\% & 0.0\%  & 7.0  & 15.3 & 126.0 \\
0.20 & 72.3\%  & 86.5\% & 0.0\%  & 11.2 & 24.8 &  86.4 \\
0.40 & 45.4\%  & 40.6\% & 3.1\%  & 15.3 & 63.5 &  37.1 \\
\hline
\end{tabular}
\end{table}

\paragraph{Discussion.}
\name is robust to moderate detection noise.
At $\sigma{=}0.05$ (5\% FN drop rate, 5\% FP injection rate), the MOTA drops by only 5.5 percentage points (pp) and all tracks remain mostly-tracked (MT\% = 95.8\%), indicating that the model's cross-camera association and temporal linking readily absorb occasional missed or spurious detections.
At $\sigma{=}0.10$ the degradation remains moderate ($-$11.7\,pp MOTA) with MT\% still at 94.8\% and zero mostly-lost tracks; the identity switch count rises to 7.0 as the increased FP rate occasionally creates phantom groups that briefly capture a track.

At $\sigma{=}0.20$, performance drops more substantially ($-$23.5\,pp MOTA, MT\% falls to 86.5\%) but no tracks are mostly-lost, demonstrating that the temporal predictor~$\rho$ and post-hoc merge successfully bridge the gaps caused by the 20\% false-negative rate.
The fragment count roughly quadruples compared to baseline (24.8 vs.\ 6.5), reflecting the increasing frequency of detection gaps that interrupt continuous tracking.

At $\sigma{=}0.40$ the system degrades severely (MOTA drops to 45.4\%, MT\% to 40.6\%), with the degradation following the same density-dependent pattern observed for camera noise (Appendix~\ref{app:robustness}).
The 8-fish scene retains 51.4\% MOTA despite the extreme noise, whereas the three 16-fish scenes fall to 29--36\% MOTA, because the high false-positive rate roughly doubles the effective number of detections per frame, amplifying the combinatorial complexity of cross-camera association in already-dense scenes.
Meanwhile, the 4- and 5-fish scenes maintain 54\% and 71\% MOTA respectively, as the low object count limits the number of plausible spurious groupings.

\section{Runtime Performance}\label{app:runtime}

Table~\ref{tab:runtime} reports the training time, inference speed, and peak GPU memory of \name.
All measurements are taken on a single NVIDIA A100 80GB PCIe GPU with batch size~1 (one frame-pair at a time).

\begin{table}[H]
\centering
\caption{\name runtime and memory profile. Training time is for the full 400-epoch schedule on SynFish (8 training scenes). Inference speed is measured per frame (all cameras), averaged over the SynFish test set.}
\label{tab:runtime}
\small
\begin{tabular}{l r}
\hline
\textbf{Metric} & \textbf{Value} \\
\hline
Parameters              & 655\,K \\
Training time           & 17h 36m \\
Inference (ms\,/\,frame) & 124.8 \\
Peak GPU memory (GB)    & 0.02 \\
\hline
\end{tabular}
\end{table}

\clearpage

\section{Extended Method}\label{app:extended_method}

This appendix contains only the implementation details explicitly referenced from the main Method section.

\subsection{Grouping, Thresholding, and Geometric Validation}\label{app:extended_grouping}

Predicted correspondence probabilities are converted into multi-view fish hypotheses through a four-stage grouping pipeline.

\paragraph{Affinity matrix.}
For every detection pair $(i,j)$ from different cameras, we construct a combined affinity score:
$$
A_{ij} = w_p\,P_{ij} + w_f\,\frac{\cos(\boldsymbol{h}_i, \boldsymbol{h}_j)+1}{2},
$$
where $P_{ij}$ is the learned correspondence probability, $\cos(\cdot,\cdot)$ denotes cosine similarity between the transformer embeddings, and the weights $w_p{=}0.6$, $w_f{=}0.4$ balance association confidence and embedding agreement. The cosine term is shifted from $[-1,1]$ to $[0,1]$ so that both terms share the same range.

\paragraph{Per-camera-pair Hungarian matching.}
For each unordered camera pair $(c_a, c_b)$, we extract the directed sub-matrix $A_{c_a, c_b}$ with rows indexed by detections from camera $c_a$ and columns indexed by detections from camera $c_b$, and solve a maximum-weight bipartite assignment using the Hungarian algorithm~\cite{kuhn1955hungarian} (equivalently, minimising $-A_{c_a,c_b}$). Because the pairwise head operates on ordered concatenations $[\boldsymbol{h}_i \| \boldsymbol{h}_j]$, the predicted matrix is generally asymmetric; the implementation therefore uses a fixed row/column camera ordering rather than symmetrising $P_{ij}$ and $P_{ji}$. Matched pairs whose affinity falls below a threshold $\tau_{\mathrm{assoc}}$ are discarded; two-camera groups require a stricter threshold to reduce false triangulations.

\paragraph{Transitive merging.}
Accepted matches are treated as edges in a graph over all $N$ detections. Connected components are computed via union-find, yielding multi-view groups that may span all $C$ cameras. Within each component, if multiple detections originate from the same camera, only the one with the highest detection confidence is retained. Groups with fewer than two unique cameras are discarded.

\paragraph{Triangulation and geometric validation.}
Each group is triangulated~\cite{hartley1997triangulation} into a candidate 3D point by solving the linear system arising from the member centroids and camera projection matrices. The candidate is validated by reprojecting the 3D point into every member camera and checking that the reprojection falls inside the corresponding bounding box. Groups that fail this containment check are rejected, ensuring that accepted hypotheses are geometrically consistent. A group embedding $\bar{\boldsymbol{h}}_g = \ell_2\text{-norm}\!\bigl(\tfrac{1}{|g|}\sum_{i\in g}\boldsymbol{h}_i\bigr)$ is computed as the $\ell_2$-normalised mean of its member transformer embeddings. The group confidence used downstream is the implementation heuristic
$$
\mathrm{conf}(g) = \min(1.0,\, 0.2 + 0.2\,n_{\mathrm{cams}}) \times
\begin{cases}
0.75, & n_{\mathrm{cams}} = 2 \\
1, & n_{\mathrm{cams}} \ge 3,
\end{cases}
$$
where $n_{\mathrm{cams}}$ is the number of unique supporting cameras. Thus, 2-view groups receive confidence $0.45$ and 3-view groups receive confidence $0.8$, making the threshold $\theta_{\mathrm{conf}}{=}0.3$ in Table~\ref{tab:hyperparams} meaningful rather than vacuous.

\subsection{Positive Mining for the Contrastive Loss}\label{app:extended_contrastive}

Within each frame, detections belonging to the same multi-view group are treated as positives, and all other detections as negatives.

\paragraph{Positive mining from grouping.}
For each group with confidence above a minimum threshold, every member detection $i$ is paired with all other members of the same group. Because groups are formed from cross-camera detections of the same fish, this yields positive pairs without any identity labels: the self-supervised grouping step provides the supervision signal.

\paragraph{Frame-level contrastive implementation.}
Given the transformer output embeddings $\{\boldsymbol{h}_i\}_{i=1}^{N}$, we first $\ell_2$-normalise to obtain unit vectors $\boldsymbol{z}_i = \boldsymbol{h}_i / \|\boldsymbol{h}_i\|$. We then compute a scaled similarity matrix $S_{ij} = \boldsymbol{z}_i^\top \boldsymbol{z}_j / \tau$, where $\tau$ is a temperature parameter. For each anchor detection $i$ with positive set $\mathcal{P}_i$ (same-group members), the loss is:
$$
\mathcal{L}_{\mathrm{ctr}}^{(i)}
= -\log \frac{\displaystyle\sum_{j\in\mathcal{P}_i}\exp(S_{ij})}
          {\displaystyle\sum_{\substack{k=1\\k\neq i}}^{N}\exp(S_{ik})},
$$
and the frame-level contrastive loss averages over all anchors: $\mathcal{L}_{\mathrm{ctr}} = \frac{1}{|\mathcal{A}|}\sum_{i\in\mathcal{A}} \mathcal{L}_{\mathrm{ctr}}^{(i)}$, where $\mathcal{A}$ is the set of detections that have at least one positive. The denominator sums over all other detections regardless of group membership, pushing different-identity embeddings apart.

\subsection{Fallback Matching and Temporal Predictor Details}\label{app:extended_temporal}

To encourage identity stability over time, \name links triangulated group hypotheses between consecutive frames and trains a temporal predictor.

\paragraph{Hungarian temporal matching.}
Given groups from frames $t$ and $t{+}1$ (each with a 3D position $\mathbf{X}$ and a group embedding $\bar{\boldsymbol{h}}$), we construct a cost matrix:
$$
\mathrm{cost}(i,j)
= \alpha\,\frac{\|\mathbf{X}_j^{(t+1)} - \mathbf{X}_i^{(t)}\|}{d_{\max}}
+ \beta\,\big(1 - \cos(\bar{\boldsymbol{h}}_i^{(t)}, \bar{\boldsymbol{h}}_j^{(t+1)})\big),
$$
where $\alpha{=}0.6$ and $\beta{=}0.4$ balance 3D proximity and embedding similarity. The assignment is solved by the Hungarian algorithm~\cite{kuhn1955hungarian}, and matched pairs are gated: a match is accepted only if the 3D distance is below a threshold $d_{\max}$ and the cosine similarity exceeds a threshold $s_{\min}$. Each accepted match receives a weight $w_{ij} = \min(\mathrm{conf}_i, \mathrm{conf}_j)$ reflecting the geometric reliability of both groups.

\paragraph{Fallback multi-frame matching.}
Groups in frame $t{+}1$ that remain unmatched may correspond to fish that were temporarily missed in frame $t$. A sliding window of past group embeddings (up to 10 frames) is maintained. For each unmatched group in $t{+}1$, we search older frames for a compatible partner, relaxing the distance gate by a factor proportional to the frame gap $g$ and reducing the similarity gate accordingly. This fallback mechanism provides training pairs even under transient occlusion, ensuring the predictor receives supervision for longer-horizon identity linking.

\paragraph{Predictor architecture and loss implementation.}
We train a temporal predictor $\rho$, a two-layer MLP, on matched groups from adjacent frames to preserve identity consistency through short occlusions. For each matched pair $(i,j)$, the predictor takes the observed group embedding $\bar{\boldsymbol{h}}_i^{(t)}$ and predicts $\hat{\bar{\boldsymbol{h}}}_j^{(t+1)} = \rho(\bar{\boldsymbol{h}}_i^{(t)})$. The loss is a weight-normalised mean squared error over all immediate and fallback matches:
$$
\mathcal{L}_{\mathrm{temp}}
= \frac{\displaystyle\sum_{(i,j)} w_{ij}\,\|\hat{\bar{\boldsymbol{h}}}_j - \bar{\boldsymbol{h}}_j^{(t+1)}\|^2}
    {\displaystyle\sum_{(i,j)} w_{ij}},
$$
where each weight $w_{ij}$ is the minimum confidence of the two matched groups, ensuring that geometrically unreliable matches contribute less to the gradient. During training the predictor always receives a real observed embedding as input; chained rollout of predicted embeddings is used only at inference for occlusion bridging.

\subsection{Tracker and Fragment Stitching Details}\label{app:extended_inference}

At inference, \name outputs full 3D trajectories using a training-faithful tracker: all identity decisions are driven by the same quantities learned during training, namely cross-view correspondence probabilities $P_{ij}$, identity-discriminative embeddings, and the temporal predictor.

\paragraph{Per-frame multi-view reconstruction.}
For each frame, we encode detections, predict $P_{ij}$ with the association transformer, form multi-view groups via Hungarian assignment on the learned affinity, and triangulate each group to obtain a 3D hypothesis with a group embedding.

\paragraph{Online temporal linking with predictor-based occlusion bridging.}
We maintain active tracks storing the last 3D position and embedding. Tracks are matched to current hypotheses with Hungarian assignment using the same distance--similarity cost and gap-dependent gate relaxation used during training-time fallback matching. When a track is missing for $g$ frames, we roll its group embedding forward with the trained predictor, $\bar{\boldsymbol{h}}^{(t+g)}\approx \rho^{g}(\bar{\boldsymbol{h}}^{(t)})$, and use this predicted embedding for matching.

\paragraph{Track-fragment stitching using learned representations.}
After the online pass, we stitch non-overlapping track fragments by comparing their boundary embeddings in the same learned space, additionally testing $\rho^{g}(\bar{\boldsymbol{h}}^{A}_{\mathrm{last}})$ against $\bar{\boldsymbol{h}}^{B}_{\mathrm{first}}$ across the gap $g$. Fragments are merged only when embedding similarity and 3D boundary proximity exceed fixed thresholds. This step introduces no new learned parameters and no extra supervision.

\paragraph{Post-processing.}
For presentation, we optionally interpolate short gaps between two already-associated observations of the same track and remove very short-lived tracks as ghosts; these operations do not create new identity correspondences and are applied to baselines for fairness.

\clearpage


\section{Extended Experiment}\label{app:extended_experiment}

This appendix contains only the supplementary material explicitly referenced from the main Results section.

\subsection{\textbf{Datasets and difficulty characterisation}}\label{app:extended_dataset}

We evaluate \name on the synthetic \textbf{SynFish} benchmark and the real-world zebrafish benchmark 3D-ZeF~\cite{pedersen2020zef}. All datasets provide calibrated cameras; SynFish additionally provides ground-truth 3D trajectories for quantitative evaluation.

\paragraph{SynFish dataset.}
The SynFish corpus contains 14 scenes with visually near-indistinguishable individuals, spanning 4--16 fish, rendered at $1920{\times}1080$ with $C{=}3$ synchronised cameras. The scenes use 7 distinct camera-rig configurations; the training set spans 6 rigs and the test set uses 4, one of which is unseen during training (Appendix~\ref{app:cam_ue5}). We split these into 8 training scenes used for self-supervised learning and 6 held-out test scenes used only for evaluation.

\paragraph{Real-world zebrafish (3D-ZeF).}
3D-ZeF uses $C{=}2$ calibrated cameras at $2704{\times}1520$ resolution with 2 or 5 fish. We use zebra1 and zebra3 for training and zebra2 and zebra4 for testing; baselines that require supervision are finetuned on the training split, while \name remains self-supervised on the corresponding split.

\paragraph{Occlusion difficulty score.}
To quantify difficulty beyond fish count, we report an occlusion score (OccScore), defined as the average, over frames and views, of the maximum pairwise 2D bounding-box IoU in that frame-view. We also report \%Overlap, the percentage of frame-view instances containing any overlap (IoU$>$0.01). Full per-scene difficulty statistics are provided in Appendix Table~\ref{tab:dataset_summary}, and camera calibration parameters are listed in Appendix~\ref{app:camera_params}.

\subsection{\textbf{Baselines and fair 3D tracking pipelines}}\label{app:extended_baselines}

No prior method solves dense multi-camera 3D tracking of visually homogeneous fish end-to-end without identity supervision, so we evaluate 13 complete 3D tracking pipelines assembled from strong components.

\begin{table}[t]
\centering
\caption{
Summary of baseline families, capability profiles, and the fair 3D tracking protocol.
\textbf{Temp}\,=\,learned temporal association;
\textbf{Spat}\,=\,learned spatial (cross-view) association;
\textbf{No Sup}\,=\,no identity supervision required;
\textbf{No Vis}\,=\,no visual appearance features used.
All methods receive the same inputs (multi-view detections and calibrated cameras) and use the same post-processing for fairness.
}
\label{tab:baseline_summary}
\small
\resizebox{\columnwidth}{!}{
\begin{tabular}{l l cccc l}
\hline
\textbf{Family} & \textbf{Methods} & \textbf{Temp} & \textbf{Spat} & \textbf{No Sup} & \textbf{No Vis} & \textbf{How 3D tracks are produced} \\
\hline
Det.\ + geometric MVA &
SAM3 / YOLO26+SAM2 + \{Hung., Greedy\} &
-- & -- & \checkmark & \checkmark &
Geometric MVA $\rightarrow$ triangulate $\rightarrow$ NN linking \\
Det.\ + geometric MVA + Kalman &
SAM3 + 3D-SORT &
-- & -- & \checkmark & \checkmark &
Geometric MVA $\rightarrow$ triangulate $\rightarrow$ Kalman tracking \\
\hline
\multirow{2}{*}{Det.\ + learned MVA} &
Self-MVA &
-- & \checkmark & \checkmark & -- &
Learned MVA $\rightarrow$ triangulate $\rightarrow$ NN linking \\
& ASNet &
-- & \checkmark & -- & -- &
Learned MVA $\rightarrow$ triangulate $\rightarrow$ NN linking \\
\hline
2D MOT $\rightarrow$ 3D &
SambaMOTR, MOTIP &
\checkmark & -- & -- & -- &
2D tracking per view $\rightarrow$ MVA at $t{=}0$ $\rightarrow$ triangulate \\
\hline
Multi-camera MOT &
ReST, MCTR &
\checkmark & \checkmark & -- & -- &
Cross-view 2D tracking $\rightarrow$ triangulate \\
\hline
Ours &
\name &
\checkmark & \checkmark & \checkmark & \checkmark &
Learned grouping + triangulation; temporal linking \\
\hline
\end{tabular}
}
\end{table}

\paragraph{Detection + geometric MVA + 3D nearest-neighbour linking.}
We pair SAM~3~\cite{sam3} and YOLO26+SAM2~\cite{yolo26,sam2} with geometric multi-view association using Hungarian~\cite{kuhn1955hungarian} or greedy matching. Costs are derived from epipolar geometry and reprojection consistency; matched detections are triangulated into 3D points and linked over time by nearest-neighbour 3D distance.

\paragraph{Detection + geometric MVA + 3D Kalman tracking.}
The 3D-SORT baseline uses the same SAM3 detections and Hungarian cross-view associations, then replaces nearest-neighbour temporal linking with the constant-velocity Kalman filtering and gated assignment described in Appendix~\ref{app:sort_baseline}.

\paragraph{Detection + learned MVA + 3D nearest-neighbour linking.}
We additionally evaluate \emph{Self-MVA}~\cite{selfmva2025} and \emph{ASNet}~\cite{messytable2020}. For each learned MVA method, we report the stronger of the SAM3 and YOLO26+SAM2 detector variants, while keeping the same nearest-neighbour 3D linking stage.

\paragraph{Single-view 2D MOT $\rightarrow$ 3D.}
We evaluate \emph{SambaMOTR}~\cite{samba2024} and \emph{MOTIP}~\cite{motip2024}, then lift their per-view 2D tracks to 3D by performing Hungarian multi-view association at the first frame and keeping that cross-view ID mapping fixed during triangulation.

\paragraph{Multi-camera MOT $\rightarrow$ triangulate.}
We evaluate \emph{ReST}~\cite{rest2023} and \emph{MCTR}~\cite{mctr2024}, whose outputs already provide cross-view-consistent 2D tracks that can be triangulated directly into 3D trajectories.

\paragraph{Finetuning / retraining protocol.}
All external components are finetuned or retrained on the SynFish and 3D-ZeF training sets using their default configurations, including any required re-identification backbone or feature extractor. \name is trained self-supervisedly on the corresponding training split using SAM3 or YOLO26+SAM2 detections.

\paragraph{Shared post-processing and fairness.}
All pipelines use the same post-processing: short-gap interpolation between already-associated observations of the same track and removal of very short-lived ghost tracks. These steps use no privileged information and create no new identity correspondences.

\subsection{\textbf{Implementation details}}\label{app:extended_impl}

\name is trained for 400 epochs with a 100-epoch warmup using only calibrated cameras and 2D observations, without identity labels or ground-truth 3D trajectories. The model comprises a geometric encoder, a 4-layer association transformer with RoPE, and a temporal predictor, totalling about \textbf{655K} trainable parameters (Appendix~\ref{app:architecture}). We use AdamW with learning rate $10^{-4}$, weight decay $10^{-4}$, gradient clipping at 5, and loss weights $\lambda_{\mathrm{asso}}{=}1.0$, $\lambda_{\mathrm{ctr}}{=}0.5$, and $\lambda_{\mathrm{temp}}{=}0.25$.

Key operational thresholds are as follows: grouping uses an association probability threshold $\theta_{\mathrm{assoc}}{=}0.6$, an embedding cosine-similarity threshold $\theta_{\mathrm{emb}}{=}0.55$, and a hybrid affinity $0.6\,P + 0.4\,\cos$; temporal linking uses a Hungarian cost $\alpha\,d_{\mathrm{3D}} + \beta\,(1{-}\cos)$ with $\alpha{=}0.6$, $\beta{=}0.4$, gated by a distance threshold $\theta_{\mathrm{dist}}{=}0.5$ and a similarity threshold $\theta_{\mathrm{sim}}{=}0.6$. Complete hyperparameters, including fallback matching and the full training schedule, are listed in Appendix Table~\ref{tab:hyperparams}. Inference uses the training-faithful tracker from Sec.~\ref{sec:inference}, with dynamic track birth/death and predictor-based embedding rollforward; further inference details are given in Appendix~\ref{app:inference}. All reported results use detections from SAM\,3 or YOLO26+SAM2 rather than ground-truth boxes, and a controlled ablation with additional synthetic detection noise is provided in Appendix Table~\ref{tab:det_robustness}.

\clearpage


\section{Proof-of-principle downstream analysis on Abnormal Behaviour Detection}\label{app:anomaly}

Detecting abnormal swimming behaviour is a critical task in aquaculture and marine ecology: sick or stressed fish often exhibit reduced speed, social isolation, and loss of schooling coordination, and early identification can prevent disease spread through the population~\cite{chong2017abnormal,li2019gan_anomaly}. However, real-world behavioural annotation is scarce and labour-intensive, motivating automated approaches that operate on 3D trajectories rather than raw video. We show that the 3D trajectories produced by \name are sufficiently accurate to support fully unsupervised anomaly detection without behavioural labels.

\paragraph{Experimental setup.}
We generate a SynFish scene with 8 fish, one of which is programmed as abnormal. The sick fish is animated with reduced swimming speed, erratic heading changes, and diminished schooling tendency, mimicking symptoms of common fish diseases~\cite{martins2012behavioural}. \name is trained self-supervisedly on this scene using the same geometric objectives and training protocol as the main paper, with no identity labels and no ground-truth 3D trajectories. We then run inference to produce 3D trajectories and apply a sliding-window unsupervised anomaly detection pipeline.

\paragraph{Trajectory feature extraction.}
We adopt a frame-level evaluation using a sliding window of $W{=}3$ frames with stride 1, yielding 359 overlapping windows. Within each window, we extract a 9-dimensional feature vector for every tracked individual: mean speed, speed standard deviation, maximum speed, total path length, straightness (net displacement / path length), mean absolute acceleration, mean distance to group centroid, standard deviation of centroid distance, and mean nearest-neighbour distance. These features capture the key behavioural signatures of sick fish: reduced locomotion, spatial isolation, and failure to coordinate with the school~\cite{papadakis2012anomaly}.

\paragraph{Anomaly detection methods.}
At each window position, we apply three unsupervised anomaly detectors to the standardised feature vectors: (i) Isolation Forest~\cite{liu2008isolation} with $n_{\mathrm{estimators}}{=}200$, (ii) Local Outlier Factor (LOF)~\cite{breunig2000lof} with $k{=}5$, and (iii) a domain-informed z-score on the two most discriminative features, mean speed and mean centroid distance, flagging individuals with combined z-score greater than $1.5$. We also report a consensus prediction via majority vote across the three methods. Short-lived ghost tracks with duration below 30\% of total frames are filtered before analysis.

\begin{figure*}[t]
\centering
\begin{minipage}[t]{0.60\textwidth}
\vspace{0pt}
\centering
\captionof{table}{Frame-level abnormal fish detection: 2D bounding-box trajectories from each individual camera view versus 3D trajectories from \name. A sliding window of $W{=}3$ frames is moved across the sequence (359 windows). We report exact-match rate, precision, recall, and F1 averaged over all windows. 2D features are extracted from bbox-centre tracks linked by nearest-neighbour association. Consensus~=~majority vote of Isolation Forest, LOF, and z-score.}
\label{tab:anomaly}
\small
\begin{tabular}{l l rrrr}
\hline
\textbf{Input} & \textbf{Method} & \textbf{Exact\%} & \textbf{Prec.} & \textbf{Rec.} & \textbf{F1} \\
\hline
\multirow{4}{*}{2D Cam\,0}
 & Iso.\ Forest    & 48.2\% & 48.2\% & 48.2\% & 48.2\% \\
 & LOF             & 45.7\% & 45.7\% & 45.7\% & 45.7\% \\
 & Z-score         & 84.7\% & 85.1\% & 85.5\% & 85.2\% \\
 & Consensus       & 52.1\% & 52.1\% & 52.1\% & 52.1\% \\
\hline
\multirow{4}{*}{2D Cam\,1}
 & Iso.\ Forest    & 51.8\% & 51.8\% & 51.8\% & 51.8\% \\
 & LOF             & 44.0\% & 44.0\% & 44.0\% & 44.0\% \\
 & Z-score         &  0.0\% &  0.0\% &  0.0\% &  0.0\% \\
 & Consensus       & 42.9\% & 42.9\% & 42.9\% & 42.9\% \\
\hline
\multirow{4}{*}{2D Cam\,2}
 & Iso.\ Forest    & 36.5\% & 36.5\% & 36.5\% & 36.5\% \\
 & LOF             & 25.9\% & 25.9\% & 25.9\% & 25.9\% \\
 & Z-score         & 12.3\% & 12.3\% & 12.3\% & 12.3\% \\
 & Consensus       & 25.9\% & 25.9\% & 25.9\% & 25.9\% \\
\hline
\multirow{4}{*}{\textbf{3D (Ours)}}
 & Iso.\ Forest    & 78.8\% & 78.8\% & 78.8\% & 78.8\% \\
 & LOF             & 80.2\% & 80.2\% & 80.2\% & 80.2\% \\
 & Z-score         & 92.2\% & 93.5\% & 94.7\% & 93.9\% \\
 & Consensus       & 80.8\% & 80.8\% & 80.8\% & 80.8\% \\
\hline
\end{tabular}
\end{minipage}
\hfill
\begin{minipage}[t]{0.34\textwidth}
\vspace{0pt}
\centering
\includegraphics[width=0.9\linewidth]{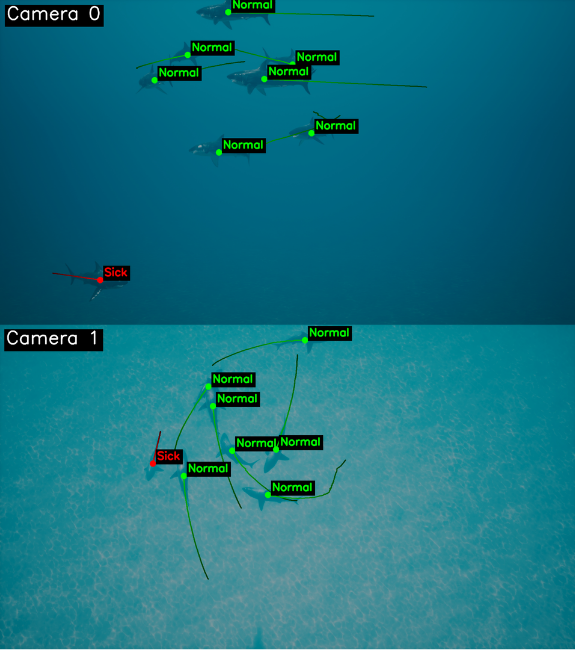}
\captionof{figure}{Anomaly detection on the 8-fish SynFish scene (1 sick fish). Trajectory-derived behavioural features---mean speed and mean centroid distance---cleanly separate the anomalous individual (red) from healthy fish (green), using only 3D trajectories predicted by \name with no behavioural labels.}
\label{fig:anomaly}
\end{minipage}
\end{figure*}

\paragraph{Results.}
Table~\ref{tab:anomaly} reports per-method detection accuracy averaged across all sliding windows, and Fig.~\ref{fig:anomaly} visualises the trajectory-derived features. The z-score detector achieves the highest window-level F1 of 93.9\%, correctly identifying the sick fish in 92.2\% of all windows. Isolation Forest and LOF perform comparably at 78.8\% and 80.2\% F1, respectively. Consensus voting across the three methods yields 80.8\% F1, showing that even with only 3 frames of observation, the anomalous individual can be detected reliably in most temporal windows.

\paragraph{Comparison with 2D-only baselines.}
To verify that 3D trajectory reconstruction genuinely contributes to anomaly detection, we repeat the same sliding-window pipeline using 2D bounding-box trajectories from each individual camera view. All 2D detections are produced by the same finetuned SAM\,3 model used by \name; the only difference is that the 2D baselines operate on per-view detections independently, whereas \name fuses detections across all three views into 3D trajectories. Because the 2D detections carry no identity labels, we construct tracks via nearest-neighbour linking of bbox centres across consecutive frames and identify the sick track by projecting the known 3D ground-truth position into each camera view.

Every camera view yields substantially lower scores than the 3D baseline: the best single-view consensus F1 is only 52.1\% (Cam\,0) versus 80.8\% for 3D trajectories, and the best single-view z-score F1 is 85.2\% (Cam\,0) versus 93.9\% for 3D. Performance varies strongly across viewpoints: Cam\,1 achieves 0.0\% z-score F1 because the sick fish's projected 2D motion in that view does not differ sufficiently from healthy individuals, and Cam\,2 drops to 25.9\% consensus F1. These results confirm that 2D bounding-box features from any single camera view are unreliable for anomaly detection, whereas calibrated 3D reconstruction recovers view-invariant behavioural features that enable consistent detection across windows.

\paragraph{Discussion.}
The sick fish exhibits roughly half the mean speed of healthy individuals and markedly higher centroid distance, producing a clear kinematic signature even within very short observation windows. The z-score detector, which directly targets the two most discriminative features, is the most robust method, while Isolation Forest and LOF are occasionally confused by natural behavioural variation among healthy individuals. Importantly, these results are obtained with a window of only 3 frames, showing that \name trajectories preserve fine-grained behavioural information sufficient for near-real-time anomaly detection. While the current evaluation uses simulated abnormalities, the same pipeline is directly applicable to real aquaculture data where calibrated multi-camera rigs are increasingly common.

\clearpage


\section{Cross-species generalisation to real-world pigeons}\label{app:pigeon}

This appendix provides the full setup and complete results for the pigeon transfer experiment summarised in Sec.~\ref{sec:eval_pigeon}.

\paragraph{Experimental setup.}
To test whether the same geometry-driven training recipe extends beyond fish, we evaluate \name on the real-world \textbf{Pigeon10} subset from 3D-POP~\cite{naik2023pop}, a four-camera benchmark for markerless 2D--3D tracking of freely moving birds. We use four sequences for training and hold out one sequence (\texttt{seq48}) for evaluation. The model is optimised on the training split with the same self-supervised objectives used throughout this paper, then tested on the held-out sequence without any bird identity labels.

\paragraph{Baselines.}
For comparison, we evaluate the same association baselines used elsewhere in the paper: geometric matching with greedy and Hungarian assignment, the supervised association model ASNet, the self-supervised appearance-based method Self-MVA, the single-view MOT baselines SambaMOTR and MOTIP, and the multi-camera trackers ReST and MCTR.

\begin{table*}[t]
\centering
\caption{Cross-species evaluation on the held-out Pigeon10 sequence \texttt{seq48} from 3D-POP. \name is trained on four real-world pigeon sequences and evaluated on the fifth. Results show that the same geometry-driven formulation transfers beyond fish to another visually similar animal collective.}
\label{tab:pigeon_results}
\resizebox{0.9\textwidth}{!}{
\begin{tabular}{l rrrrrrrrr}
\hline
\textbf{Method} & \textbf{MOTA\,$\uparrow$} & \textbf{Recall\,$\uparrow$} & \textbf{Prec.\,$\uparrow$} & \textbf{IDF1\,$\uparrow$} & \textbf{IDSW\,$\downarrow$} & \textbf{Frag\,$\downarrow$} & \textbf{MT\,$\uparrow$} & \textbf{PT\,$\downarrow$} & \textbf{ML\,$\downarrow$} \\
\hline
Greedy                           & 86.0\% & 92.6\% & 95.2\% & 63.3\% & 60 & 36 &  8 & 2 & 0 \\
Hungarian                        & 91.5\% & 93.9\% & 97.6\% & 92.0\% & 3 &  6 & \textbf{10} & \textbf{0} & 0 \\
ASNet~\cite{messytable2020}     & 85.6\% & 91.6\% & 95.3\% & 63.0\% & 46 & 40 &  9 & 1 & 0 \\
Self-MVA~\cite{selfmva2025}     & 68.7\% & 78.4\% & 90.7\% & 47.8\% & 50 & 61 &  6 & 4 & 0 \\
ReST~\cite{rest2023}            & 88.3\% & 91.5\% & \textbf{98.5\%} & 58.0\% & 56 & 35 &  8 & 2 & 0 \\
MCTR~\cite{mctr2024}            & 28.3\% & 58.2\% & 67.7\% & 35.7\% & 66 & 87 &  1 & 7 & 2 \\
SambaMOTR~\cite{samba2024}      & 74.0\% & 79.1\% & 94.0\% & 85.8\% &  5 & 36 &  6 & 3 & 1 \\
MOTIP~\cite{motip2024}          & 90.4\% & 94.1\% & 96.3\% & \textbf{95.1\%} & \textbf{1} &  6 &  9 & 1 & 0 \\
\name                           & \textbf{91.9\%} & \textbf{94.5\%} & 97.4\% & 91.9\% & 3 & \textbf{4} & 9 & 1 & 0 \\
\hline
\end{tabular}
}
\end{table*}

\paragraph{Discussion.}
These results suggest that the core assumption behind \name is not fish-specific. When individuals are visually similar and multi-view geometry is available, the same self-supervised formulation can transfer to other animal species---here, freely moving pigeons---without requiring any appearance-based identity supervision. In particular, \name achieves the best MOTA (91.9\%) and recall (94.5\%), while reducing fragmentation from 6 to 4 relative to the strongest competing baselines. MOTIP attains the strongest IDF1 (95.1\%) and the lowest IDSW count (1), but \name remains the top overall method on detection-and-tracking accuracy.

\section{3D-SORT Kinematic Baseline}
\label{app:sort_baseline}

To compare the learned temporal predictor against standard kinematic filtering, we implement a 3D adaptation of SORT~\cite{bewley2016simple}. Per-frame cross-view association is identical to the SAM3 + Hungarian baseline: detections are matched using calibrated triangulation and pairwise reprojection cost. Each resulting 3D track is then represented by a constant-velocity Kalman filter. The filter predicts the next 3D position, and unmatched incoming groups are assigned to predicted positions by gated nearest-neighbour matching. Unmatched tracks are maintained for up to 30 frames on SynFish and 3 frames on 3D-ZeF.

\begin{table}[H]
\centering
\caption{Comparison with the 3D-SORT kinematic baseline.}
\label{tab:sort_baseline}
\small
\begin{tabular}{llrrrrrr}
\toprule
Dataset & Method & MOTA $\uparrow$ & MT\% $\uparrow$ & ML\% $\downarrow$ & IDSw $\downarrow$ & Frag $\downarrow$ & MTBF $\uparrow$ \\
\midrule
SynFish & SAM3 + Hungarian & 83.5\% & 71.9\% & 0.0\% & 7.5 & 12.8 & 183.8 \\
SynFish & 3D-SORT & 87.7\% & 77.1\% & 0.0\% & 4.0 & 9.0 & 209.4 \\
SynFish & \name (SAM3) & \textbf{95.8\%} & \textbf{95.8\%} & \textbf{0.0\%} & \textbf{0.0} & \textbf{6.5} & \textbf{266.3} \\
\midrule
3D-ZeF & Hungarian & 77.4\% & 90.0\% & 0.0\% & 6.0 & 73.0 & 42.5 \\
3D-ZeF & 3D-SORT & 75.6\% & 90.0\% & 0.0\% & 13.5 & 65.5 & 43.4 \\
3D-ZeF & \name & \textbf{81.1\%} & \textbf{90.0\%} & \textbf{0.0\%} & \textbf{4.0} & \textbf{30.5} & \textbf{93.6} \\
\bottomrule
\end{tabular}
\end{table}

The Kalman filter improves the geometric baseline on SynFish by bridging short gaps under approximately smooth motion, but remains 8.1 percentage points below \name (SAM3). On 3D-ZeF, constant-velocity prediction reduces fragmentation but increases identity switches and lowers MOTA. When visually identical fish cross or turn unpredictably, a predicted position may become closer to a different fish, causing an identity swap. In contrast, \name predicts in a contextual embedding space that represents each fish relative to the full scene rather than relying only on individual linear kinematics.

\section{Density and Computational Scaling}
\label{app:scalability}

The original SynFish benchmark contains at most 16 fish. We additionally render one 30-fish test scene and evaluate the unchanged SAM3-based model and direct geometric baseline:

\begin{table}[H]
\centering
\caption{Single-scene 30-fish SynFish stress test.}
\label{tab:synfish_30}
\small
\begin{tabular}{lrrrrr}
\toprule
Method & MOTA $\uparrow$ & MT\% $\uparrow$ & IDSw $\downarrow$ & Frag $\downarrow$ & MTBF $\uparrow$ \\
\midrule
\name (SAM3) & \textbf{70.2\%} & \textbf{63.3\%} & \textbf{8} & \textbf{22} & \textbf{143.7} \\
SAM3 + Hungarian & 29.9\% & 6.7\% & 40 & 54 & 36.0 \\
\bottomrule
\end{tabular}
\end{table}

Let $N$ be the total number of detections from all cameras in a frame and $d$ the embedding dimension. Global self-attention scales as $O(N^2d)$ and stores $N^2$ pairwise attention entries. Assuming three detections per fish across the three-camera SynFish setup, the approximate scaling is:

\begin{table}[H]
\centering
\caption{Approximate attention scaling with fish count for the three-camera SynFish configuration.}
\label{tab:attention_scaling}
\small
\begin{tabular}{rrr}
\toprule
Fish & Detections per frame ($N$) & Attention entries ($N^2$) \\
\midrule
16 & 48 & 2,304 \\
30 & 90 & 8,100 \\
32 & 96 & 9,216 \\
64 & 192 & 36,864 \\
100 & 300 & 90,000 \\
\bottomrule
\end{tabular}
\end{table}

Increasing from 16 to 30 fish therefore increases the attention matrix by approximately $3.5\times$, but 8,100 entries remain modest. Computation may become relevant for substantially larger schools, where sparse or hierarchical attention would be appropriate. At the evaluated densities, the more immediate limitation is geometric: higher density increases mutual occlusion, missed detections, and the number of plausible cross-view correspondences. If too few fish remain visible in at least two cameras, reliable triangulation becomes impossible regardless of the association model. The 30-fish result in Table~\ref{tab:synfish_30} is consequently a synthetic single-scene stress test, not a universal capacity threshold or evidence of real high-density generalisation.

\section{Pseudo-Label Noise Audit}
\label{app:pseudolabel_audit}

We directly audit geometry-derived pseudo-positive pairs against SynFish ground-truth identities. Ground truth is used only for this diagnostic evaluation and is never provided during training. A pseudo-positive is counted as correct when both detections belong to the same ground-truth fish identity.

\begin{table}[H]
\centering
\caption{Pairwise quality of geometry-derived pseudo-positive associations as density increases.}
\label{tab:pseudolabel_audit}
\small
\begin{tabular}{rrr}
\toprule
Fish & Precision $\uparrow$ & Recall $\uparrow$ \\
\midrule
4 & 70.8\% & 100.0\% \\
16 & 25.3\% & 100.0\% \\
30 & 17.0\% & 100.0\% \\
\bottomrule
\end{tabular}
\end{table}

Precision falls substantially as geometrically plausible incorrect pairs become more common, while recall remains complete in the evaluated scenes. Thus, we do not claim that pseudo-labels are clean or intrinsically stable under density. Instead, they provide noisy, high-coverage supervision aggregated across detections, views, and frames. The learned objectives exploit this repeated signal, while direct geometric matching treats each ambiguous association as a final decision.

\section{Training-Time Calibration Noise}
\label{app:training_calibration}

Appendix~\ref{app:robustness} perturbs calibration only at inference while training on clean geometry. To separately evaluate calibration uncertainty during pseudo-label generation and geometric feature construction, we conduct a controlled 3D-ZeF experiment in which the training camera extrinsics receive Gaussian rotation and translation perturbations with $\sigma_R{=}\sigma_t{=}0.1$. We regenerate all geometry-derived supervision under the perturbed calibration, train \name on these data, and evaluate using the original clean calibration and ground truth.

\begin{table}[H]
\centering
\caption{3D-ZeF performance after training with perturbed calibration and evaluating with clean calibration.}
\label{tab:training_calibration}
\small
\begin{tabular}{lrrrrrr}
\toprule
Training calibration & MOTA $\uparrow$ & MT\% $\uparrow$ & ML\% $\downarrow$ & IDSw $\downarrow$ & Frag $\downarrow$ & MTBF $\uparrow$ \\
\midrule
Clean & 81.1\% & 90.0\% & 0.0\% & 4.0 & 30.5 & 93.6 \\
Noisy, $\sigma{=}0.1$ & 66.8\% & 60.0\% & 0.0\% & 13.5 & 35.5 & 59.0 \\
\bottomrule
\end{tabular}
\end{table}

Training-time calibration noise causes a clear degradation relative to clean training, so this experiment does not establish calibration invariance. Nevertheless, performance does not collapse and no trajectories become mostly lost. The result supports the narrower conclusion that useful tracking performance remains under this perturbation while confirming calibration quality as a practical limitation.

\end{document}